\documentclass{article}

\usepackage{microtype}
\usepackage{graphicx}
\usepackage{booktabs} 

\usepackage[pagebackref=true,breaklinks=true,letterpaper=true,colorlinks,bookmarks=false]{hyperref}

\usepackage[accepted]{icml2018}
\usepackage{xcolor}
\usepackage{array}
\newcommand{\modelName}{COCO-GAN }
\newcommand{\modelNamePunc}{COCO-GAN}
\newcommand{\expectation}{\mathop{\mathbb{E}}}

\newcommand\blfootnote[1]{%
  \begingroup
  \renewcommand\thefootnote{}\footnote{#1}%
  \addtocounter{footnote}{-1}%
  \endgroup
}

\definecolor{tfboardRedDef}{rgb}{0.8, 0.2, 0.0627}
\definecolor{tfboardDeepBlueDef}{rgb}{0.1882, 0.4118, 0.6784}
\definecolor{tfboardLightBlueDef}{rgb}{0.2, 0.7333, 0.9333}
\definecolor{tfboardPinkDef}{rgb}{0.9333, 0.1961, 0.4667}
\definecolor{tfboardOrangeDef}{rgb}{0.9608, 0.3922, 0.2235}

\usepackage{ulem}
\definecolor{crimson}{rgb}{0.86, 0.08, 0.24}
\definecolor{green}{rgb}{0, 0.5, 0.25}
\definecolor{purple}{rgb}{0.75, 0, 1}
\definecolor{orange}{rgb}{1, 0.5, 0.25}
\definecolor{yellow}{rgb}{1, 1, 0}
\definecolor{new_blue}{rgb}{0, 0.5, 1}
\ifx \submission \undefined

\else

\fi

\newcolumntype{C}[1]{>{\centering\arraybackslash}m{#1}}
\newcolumntype{R}[1]{>{\raggedleft\arraybackslash}m{#1}}
\newcolumntype{L}[1]{>{\raggedright\arraybackslash}m{#1}}

\usepackage{xspace}
\makeatletter
\DeclareRobustCommand\onedot{\futurelet\@let@token\@onedot}
\def\@onedot{\ifx\@let@token.\else.\null\fi\xspace}

\def\eg{\emph{e.g}\onedot} 
\def\ie{\emph{i.e}\onedot}

\makeatother

\usepackage{amsmath,amsfonts,bm}

\def\eqref#1{equation~\ref{#1}}

\def\1{\bm{1}}

\DeclareMathAlphabet{\mathsfit}{\encodingdefault}{\sfdefault}{m}{sl}
\SetMathAlphabet{\mathsfit}{bold}{\encodingdefault}{\sfdefault}{bx}{n}

\usepackage{array}
\usepackage{subfig}
\usepackage{float}
\usepackage{makecell}
\usepackage{wrapfig}
\usepackage{booktabs}
\usepackage{multirow}
\usepackage{multicol}
\usepackage{anyfontsize}
\usepackage{adjustbox}
\usepackage{tabularx}

\usepackage[titletoc,toc,title]{appendix}

\begin{document}





\newcommand{\authorcell}[4]{
    \makecell{
        {\large #1} \\
        {\small #2} \\
        {\small \texttt{#4}}
    }
}

\twocolumn[{%
\icmltitle{COCO-GAN: Generation by Parts via Conditional Coordinating}
\renewcommand\twocolumn[1][]{#1}%
  \vspace{-2mm} 
  \centering
    {\large
    \begin{tabular}{c c c} 
        \authorcell{Chieh Hubert Lin}{National Tsing Hua University}{Hsinchu, Taiwan}{hubert052702@gmail.com} & 
        \authorcell{Chia-Che Chang}{National Tsing Hua University}{Hsinchu, Taiwan}{chang810249@gmail.com} & 
        \authorcell{Yu-Sheng Chen}{National Taiwan University}{Taipei, Taiwan}{\makecell{nothinglo@\\cmlab.csie.ntu.edu.tw}} \\ [2em]
        \authorcell{Da-Cheng Juan}{Google AI}{Mountain View, CA, USA}{dacheng@google.com} & 
        \authorcell{Wei Wei}{Google AI}{Mountain View, CA, USA}{wewei@google.com} & 
        \authorcell{Hwann-Tzong Chen}{National Tsing Hua University}{Hsinchu, Taiwan}{htchen@cs.nthu.edu.tw} \\ [1em]
    \end{tabular}} \\
    \icmlkeywords{Machine Learning, Deep Learning, Generative Model, Generative Adversarial Networks, GANs, Conditional Coordinate}
    \vskip 0.3in
}]


\begin{abstract}





Humans can only interact with part of the surrounding environment due to biological restrictions. Therefore, we learn to reason the spatial relationships across a series of observations to piece together the surrounding environment. Inspired by such behavior and the fact that machines also have computational constraints, we propose \underline{CO}nditional \underline{CO}ordinate GAN (COCO-GAN) of which the generator generates images by parts based on their spatial coordinates as the condition. On the other hand, the discriminator learns to justify realism across multiple assembled patches by global coherence, local appearance, and edge-crossing continuity. Despite the full images are never generated during training, we show that COCO-GAN can produce \textbf{state-of-the-art-quality} full images during inference. We further demonstrate a variety of novel applications enabled by teaching the network to be aware of coordinates. First, we perform extrapolation to the learned coordinate manifold and generate off-the-boundary patches. Combining with the originally generated full image, COCO-GAN can produce images that are larger than training samples, which we called ``beyond-boundary generation''. We then showcase panorama generation within a cylindrical coordinate system that inherently preserves horizontally cyclic topology. On the computation side, COCO-GAN has a built-in divide-and-conquer paradigm that reduces memory requisition during training and inference, provides high-parallelism, and can generate parts of images on-demand. 

\end{abstract}
\section{Introduction}

    The human perception has only partial access to the surrounding environment due to biological restrictions (such as the limited acuity area of the fovea), and therefore humans infer the whole environment by ``assembling'' few local views obtained from their eyesight. This recognition can be done partially because humans are able to associate the spatial coordination of these local views with the environment (where they are situated in), then correctly assemble these local views, and recognize the whole environment. Currently, most of the computational vision models assume to have access to full images as inputs for down-streaming tasks, which sometimes may become a computational bottleneck of modern vision models when dealing with large field-of-view images. This limitation piques our interest and raises an intriguing question: ``\textit{is it possible to train generative models to be aware of coordinate system for generating local views (\ie parts of the image) that can be assembled into a globally coherent image?}'' 
    \blfootnote{Due to file size limit, all images are compressed, please access the full resolution pdf from: {\color{blue}\urlstyle{same}\url{http://bit.ly/COCO-GAN-full}}}
    
    
    \begin{figure}[t]
        \centering
        \includegraphics[width=\linewidth]{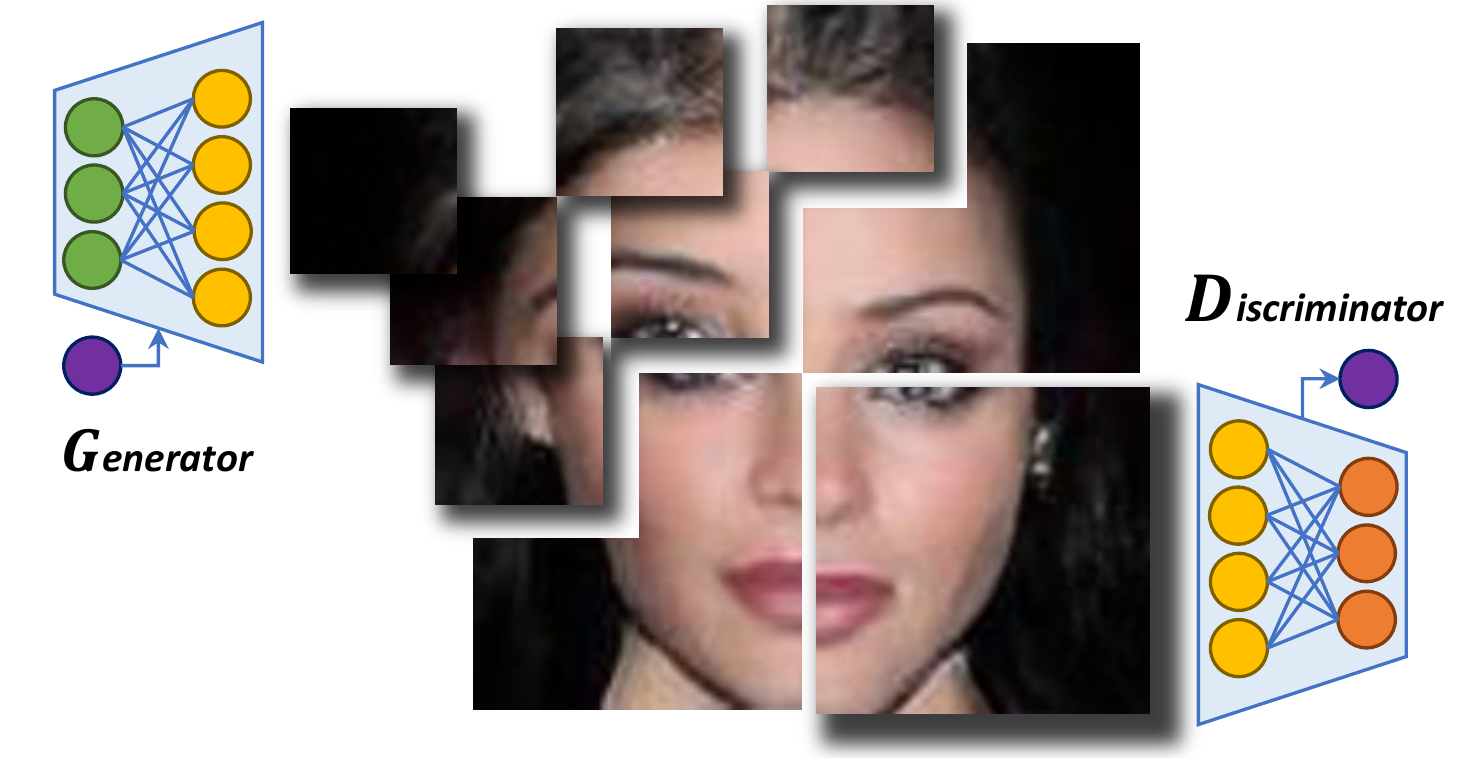}
        \caption{\modelName generates and discriminates only parts of the full image via conditional coordinating. Despite the full images are never generated during training, the generator can still produce full images that are visually indistinguishable to standard GAN samples during inference.}
        \label{fig:first-page-example-v2}
        \vspace{-1em}
    \end{figure} 
    
    \begin{figure*}[t]
        \centering
        \includegraphics[width=\linewidth]{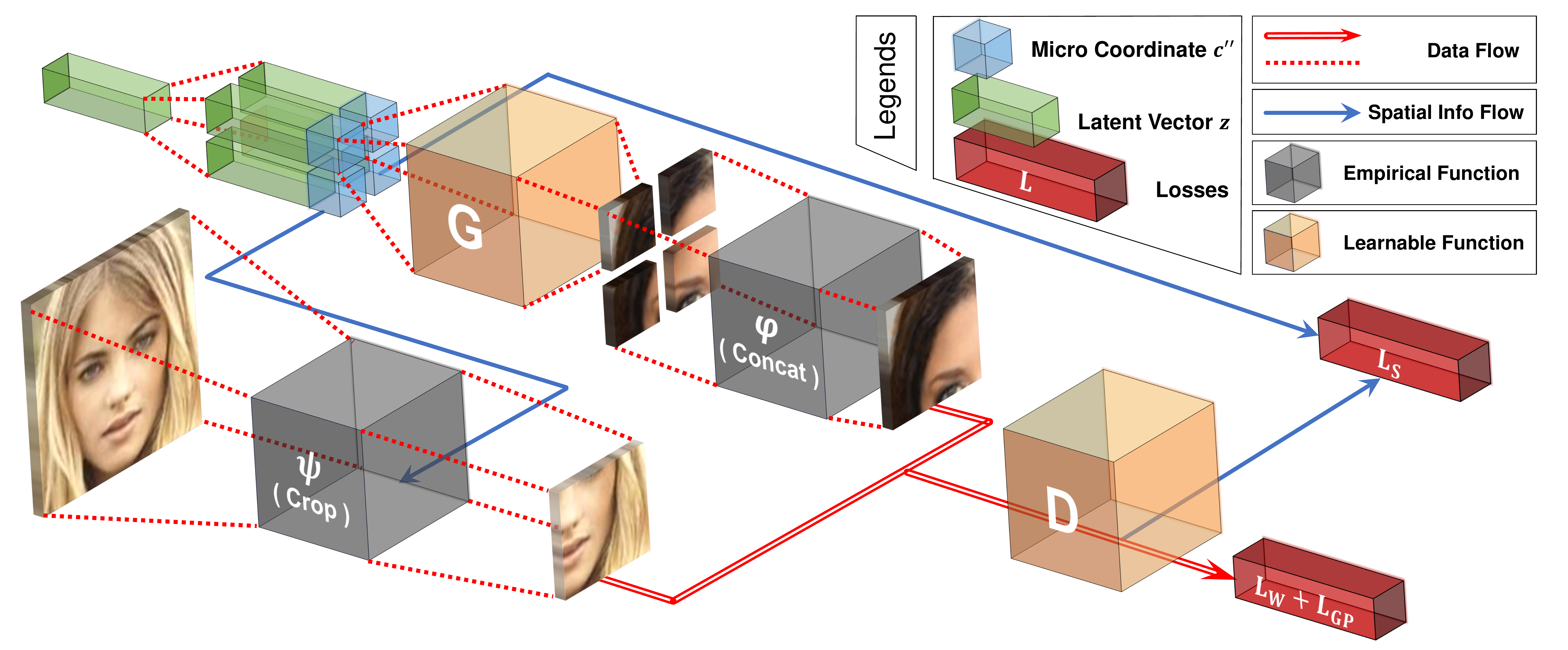}
        \vspace{-1.5em}
        \caption{An overview of \modelName training. The latent vectors are duplicated multiple times, concatenated with micro coordinates, and feed to the generator to generate micro patches. Then we concatenate multiple micro patches to form a larger macro patch. The discriminator learns to discriminate between real and fake macro patches and an auxiliary task predicting the coordinate of the macro patch. Note that the full images are only generated in the testing phase (Appendix~\ref{fig:model-testing-phase}).}
        \label{fig:model-architecture}
        \vspace{-1em}
    \end{figure*}
    
    Conventional GANs~\cite{GANs} target at learning a generator that models a mapping from a prior latent distribution (normally a unit Gaussian) to the real data distribution. To achieve generating high-quality images by parts, we introduce coordinate systems within an image and divide image generation into separated parallel sub-procedures. Our framework, named \underline{CO}nditional \underline{CO}ordinate GAN (COCO-GAN), aims at learning a coordinate manifold that is orthogonal to the latent distribution manifold. After a latent vector is sampled, the generator conditions on each spatial coordinate and generate patches at each corresponding spatial position. On the other hand, the discriminator learns to judge whether adjacent patches are structurally sound, visually homogeneous, and continuous across the edges between multiple patches. Figure~\ref{fig:first-page-example-v2} depicts the high-level idea.
    
    
    We perform a series of experiments that set the generator to generate patches under different configurations. The results show that \modelName can achieve \emph{state-of-the-art} generation quality in multiple setups with ``Fréchet Inception Distance'' (FID)~\cite{fid} score measurement. Furthermore, to our surprise, even if the generated patch sizes are set to as small as $4\times4$ pixels, the full images that are composed by \textit{\textbf{1024}} separately generated patches can still consistently form complete and plausible human faces. To further demonstrate the generator indeed learns the coordinate manifold, we perform an extrapolation experiment on the coordinate condition. Interestingly, the generator is able to generate novel contents that are never explicitly presented in the real data. We show that \modelName can produce $384\times384$ images that are larger than the $256\times256$ real training samples. We call such a procedure ``beyond-boundary generation''; all the samples created through this procedure are \textbf{\emph{guaranteed}} to be novel samples, which is a powerful example of artificial creativity.
    
    We then investigate another series of novel applications and merits brought about by teaching the network to be aware of the coordinates. The first is panorama generation. To preserve the native horizontally-cyclic topology of panoramic images, we apply cylindrical coordinate to \modelName training process and show that the generated samples are indeed horizontally cyclic. Next, we demonstrate that the ``image generation by parts'' schema is highly parallelable and saves a significant amount of memory for both training and inference. Furthermore, as the generation procedures of patches are disjoint, \modelName inherently supports generation on-demand, which particularly fits applications for computation-restricted environments, such as mobile and virtual reality. Last but not the least, we show that by adding an extra prediction branch that reconstructs latent vectors, \modelName can generate an entire image with respect to a patch of real image as guidance, which we call ``patch-guided generation''. 
    
    \modelName unveils the potential of generating high-quality images with conditional coordinating. This property enables a wide range of new applications, and can further be used by other tasks with encoding-decoding schema. With the ``generation by parts'' property, \modelName is highly parallelable and intrinsically inherits the classic divide-and-conquer design paradigm, which facilitates future research toward large field-of-view data generation.

\section{\modelName}

    \paragraph{Overview.} \modelName consists of two networks (a generator $G$ and a discriminator $D$), two coordinate systems (a finer-grained micro coordinate system for $G$ and a coarser-grained macro coordinate system for $D$), and images of three sizes: full images (real: $x$, generated: $s$), macro patches (real: $x'$, generated: $s'$) and micro patches (generated: $s''$). \blfootnote{We list all the used symbols in Appendix~\ref{appendix:symbols}.}
    
    The generator of \modelName is a conditional model that generates micro patches with $s'' = G(z, c'')$, where $z$ is a latent vector and $c''$ is a micro coordinate condition designating the spatial location of $s''$ to be generated. The final goal of $G$ is to generate realistic and \textit{seamless} full images by assembling a set of $s''$ altogether with a merging function $\varphi$. In practice, we find that setting $\varphi$ as a concatenation function without overlapping is sufficient for \modelName to synthesize high-quality images. Note that the size of the micro patches and $\varphi$ also imply a cropping transformation $\psi$, cropping out a macro patch $x'$ from a real image $x$, which is used to sample real macro patches for training $D$.
    
    In the above setting, the seams between consecutive patches become the major obstacle of full image realism. To mitigate this issue, we train the discriminator with larger macro patches that are assembled with multiple micro patches. Such a design aims to introduce the continuity and coherence of multiple consecutive or nearby micro patches into the consideration of adversarial loss. In order to fool the discriminator, the generator has to close the gap at the boundaries between the generated patches.  
    
    
    \modelName is trained with three loss terms: patch Wasserstein loss $L_W$, patch gradient penalty loss $L_{GP}$, and spatial consistency loss $L_S$. For $L_W$ and $L_{GP}$, compared with conventional GANs that use full images $x$ for both $G$ and $D$ training, \modelName only cooperates with macro patches and micro patches. Meanwhile, the spatial consistency loss $L_S$ is an ACGAN-like~\cite{ACGAN} loss function. Depending on the design of $\varphi$, we can calculate macro coordinate $c'$ for the macro patches $x'$. $L_S$ aims at minimizing the distance loss between the real macro coordinate $c'$ and the discriminator-estimated macro coordinate $\hat{c}'$. The loss functions of \modelName are
    \begin{equation}
        \begin{cases}
            L_W + \lambda L_{GP} + \alpha L_S, & \text{for the discriminator} \, D, \\
            - L_W + \alpha L_S,        & \text{for the generator} \, G.
        \end{cases}
    \end{equation}


    \paragraph{Spatial coordinate system.} 
    We start with designing the two spatial coordinate systems, a \emph{micro} coordinate system for the generator $G$ and a \emph{macro} coordinate system for the discriminator $D$. Depending on the design of the aforementioned merging function $\varphi$, each macro coordinate $c'_{(i,j)}$ is associated with a matrix of micro coordinates: ${\bm{C}''}_{(i,j)} = \big[ c''_{(i:i+N,j:j+M)} \big]$, whose complete form is
    $$ \scriptsize 
    \bm{C}''_{(i,j)} =
    \begin{bmatrix}
        c''_{(i,j)}     & c''_{(i,j+1)}     & \dots  & c''_{(i,j+M-1)} \\
        c''_{(i+1,j)}   & c''_{(i+1,j+1)}   & \dots  & c''_{(i+1,j+M-1)} \\
        \vdots          & \vdots            & \ddots & \vdots \\
        c''_{(i+N-1,j)} & c''_{(i+N-1,j+1)} & \dots  & c''_{(i+N-1,j+M-1)}
    \end{bmatrix}.
    $$
    During \modelName training, we uniformly sample all combinations of $\bm{C}''_{(i,j)}$. The generator $G$ conditions on each micro coordinate $c''_{(i,j)}$, and learns to accordingly produce micro patches $s''_{(i,j)}$ by $G(z,c''_{(i,j)})$. The matrix of generated micro patches
    $\bm{S}''_{(i,j)} = G(z,\bm{C}''_{(i,j)})$ are produced \emph{\textbf{independently}} while sharing the same latent vector $z$ across the micro coordinate matrix.
    
    The design principle of the $\bm{C}''_{(i,j)}$ construction is that, the accordingly generated micro patches $\bm{S}''_{(i,j)}$ should be spatially close to each other. Then the micro patches are merged by the merging function $\varphi$ to form a complete macro patch $s'_{(i,j)} = \varphi(\bm{S}''_{(i,j)})$ as a coarser partial-view of the imagery full-scene. Meanwhile, we assign $s'_{(i,j)}$ with a new macro coordinate $c'_{(i,j)}$ under the macro coordinate system with respect to $\bm{C}''_{(i,j)}$. On the real data side, we directly sample macro coordinates $c'_{(i,j)}$, then produce real macro patches $x'_{(i,j)} = \psi(x,c'_{(i,j)})$ with the cropping function $\psi$. Note that the design choice of the micro coordinates $\bm{C}''_{(i,j)}$ is also correlated with the topological characteristic of the micro/macro coordinate systems (for instance, the cylindrical coordinate system for panoramas used in Section~\ref{exp:panorama-generation}).
    
    In Figure~\ref{fig:model-architecture}, we illustrate one of the most straightforward designs for the above heuristic functions that we have adopted throughout our experiments. The micro patches are always a neighbor of each other and can be directly combined into a square-shaped macro patch using $\varphi$.
    We observe that setting $\varphi$ to be a concatenation function is sufficient for $G$ to learn smoothly, and eventually to produce seamless and high-quality images.

    During the testing phase, depending on the design of the micro coordinate system, we can infer a corresponding spatial coordinate matrix $\bm{C''}_{\textbf{\textit{full}}}$. Such a matrix is used to independently produce all the micro patches required for constituting the full image. 

    \begin{figure*}[t]
        \centering
        \begin{minipage}[t]{0.485\linewidth}
            \subfloat[CelebA (N2,M2,S32) (full image: 128$\times$128).]{  
                \includegraphics[width=\linewidth]{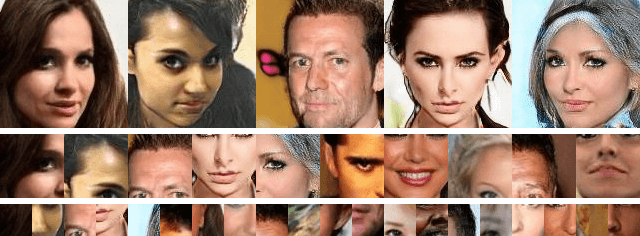}
                \label{fig:full-image-generation-celeba}
            } \\[-0.8em]
        \end{minipage}
        \hfill
        \begin{minipage}[t]{0.485\linewidth}
            \subfloat[LSUN bedroom (N2,M2,S64) (full image: 256$\times$256).]{
                \includegraphics[width=\linewidth]{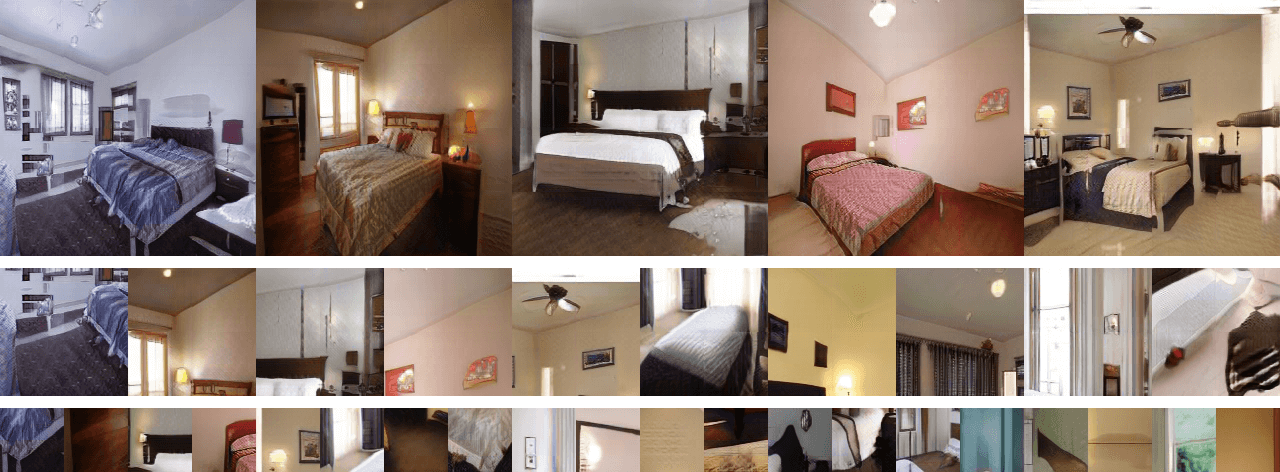}
            } \\[-0.8em]
        \end{minipage}
        \caption{\modelName generates visually smooth and globally coherent full images without any post-processing. The three rows from top to bottom show: (a) the generated full images, (b) macro patches, and (c) micro patches. For the first five columns, each column uses the same latent vector, \eg, the leftmost full image (first row), the leftmost micro patch (second row), and the leftmost micro patch (third row) share the same latent vector. Note that the columns are not aligned due to different sizes. More results can be found in the Appendix~\ref{appendix:more-full-images}.}
        \label{fig:generation-basic}
        \vspace{-0.5em}
    \end{figure*}
    

    \paragraph{Loss functions.} The patch Wasserstein loss $L_W$ is a macro-patch-level Wasserstein distance loss similar to Wasserstein-GAN~\cite{WGAN} loss. It forces the discriminator to distinguish between the real macro patches $x'$ and fake macro patches $s'$, and on the other hand, encourages the generator to confuse the discriminator with seemingly realistic micro patches $s''$. Its complete form is
    \begin{equation} 
        L_W = \displaystyle \expectation_{x,c'} \left[ \, D(\psi(x,c')) \, \right] - \expectation_{z,\bm{C}''} \left[ \, D(\varphi(G(z,\bm{C}'')) \, \right] \,. 
    \end{equation}
    Again, note that $G(z,\bm{C}'')$ represents that the micro patches are generated through independent processes. We also apply Gradient Penalty~\cite{WGAN-GP} to the macro patches discrimination:
    \begin{equation} 
    L_{GP} = \displaystyle \expectation_{\hat{s}'} \left[ ( \| \nabla_{\hat{s}'} D(\hat{s}') \|_2 - 1 )^2 \right]  \, ,
    \end{equation}
    where $\hat{s}' = \epsilon \, s' + (1-\epsilon) \, x'$ is calculated between randomly paired $s'$ and $x'$ with a random number $\epsilon \in \big[0, 1\big]$.

    Finally, the spatial consistency loss $L_S$ is similar to ACGAN loss~\cite{ACGAN}. The discriminator is equipped with an auxiliary prediction head $A$, which aims to estimate the macro coordinate of a given macro patch with $A(x')$. A slight difference is that both $c''$ and $c'$ have relatively more continuous values than the discrete setting of ACGAN. As a result, we apply a distance measurement loss for $L_S$, which is an $L_2$-loss. It aims to train $G$ to generate corresponding micro patches by $G(z,c'')$ with respect to the given spatial condition $c''$. The spatial consistency loss is
    \begin{equation}
        L_S = \displaystyle \expectation_{c'} \left[ \| c' - A(x') \|_2 \right] \, .
    \end{equation}

\section{Experiments}

    \subsection{Quality of Generation by Parts}
    
        \label{exp:generation-image-quality}
    
        We start with validating \modelName on two common GANs testbeds: CelebA~\cite{CelebA} and LSUN~\cite{LSUN} (bedroom). To verify that \modelName can learn to generate the full image without the access to the full image, we first conduct a basic setting for both datasets in which the macro patch edge length (CelebA: $64\times 64$, LSUN: $128\times 128$) is 1/2 of the full image and the micro patch edge length (CelebA: $32\times 32$, LSUN: $64\times 64$) is 1/2 of the macro patch. We denote the above cases as CelebA (N2,M2,S32) and LSUN (N2,M2,S32), where N2 and M2 represent that a macro patch is composed of $2\times2$ micro patches, and S32 means each of the micro patches is $32\times32$ pixels.
        Our results in Figure~\ref{fig:generation-basic} show that \modelName generates high-quality images in the settings that the micro patch size is 1/16 of the full image.
        
\begin{figure}[t!]
            \subfloat[CelebA (N4,M4,S16) (full image: 128$\times$128, FID: 9.99).]{  
                \includegraphics[width=\linewidth]{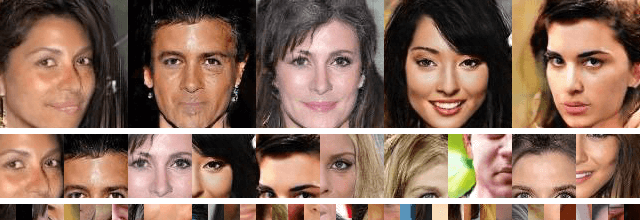}
            } \\[-0.8em]
            \subfloat[CelebA (N8,M8,S8) (full image: 128$\times$128, FID: 15.99).]{  
                \includegraphics[width=\linewidth]{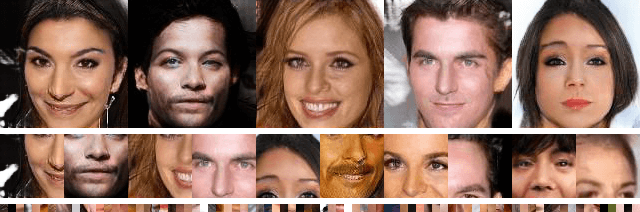}
            } \\[-0.8em]
            \subfloat[CelebA (N16,M16,S4) (full image: 128$\times$128, FID: 23.90).]{  
                \includegraphics[width=\linewidth]{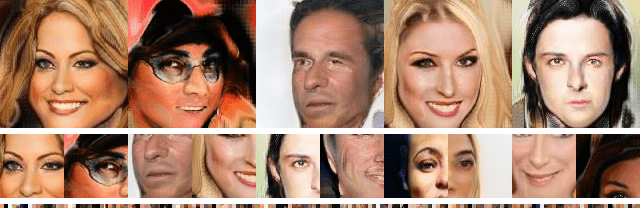}
            } 
            \caption{Various sizes of micro patches (from $16\times16$ to $4\times4$, even smaller than any human face organs) consistently generate visually smooth and globally coherent full images. Each sub-figure consists of three rows, from top to bottom: full images, macro patches, and micro patches. For the first five columns, each column uses the same latent vector (similar with Figure~\ref{fig:generation-basic}). Better to view in high-resolution since the micro patches are very small. More generation results are available in the Appendix~\ref{appendix:more-full-images}.}
            \label{fig:generation-crazy}
            \vspace{-0.5em}
        \end{figure}

        To further show that \modelName can learn more fine-grained and tiny micro patches under the same macro patch size setting, we sweep through the resolution of micro patch from $32\times32$, $16\times16$, $8\times8$, $4\times4$, labelled as (N2,M2,S32), (N4,M4,S16), (N8,M8,S8) and (N16,M16,S4), respectively. The results shown in Figure~\ref{fig:generation-crazy} suggest that \modelName can learn coordinate information and generate images by parts even with extremely tiny $4\times4$ pixels micro patch. 
        
        \label{exp:fid-score}
        
        We report Fréchet Inception Distance (FID)~\cite{fid} in Table~\ref{table:fid-score} comparing with state-of-the-art GANs. Without additional hyper-parameter tuning, the quantitative results show that \modelName is competitive with other state-of-the-art GANs. In Appendix~\ref{appendix:indicator-curves}, we also provide Wasserstein distance and FID score through time as training indicators. The curves suggest that \modelName is stable during training.
        
        \begin{table}[t]
            \centering
            \footnotesize
            \setlength\tabcolsep{2.5pt}
            \begin{tabular}{cccccc}
                \toprule \\ [-0.9em]
                Dataset & \makecell{CelebA \\ 64$\times$64} & \makecell{CelebA \\ 128$\times$128} & \makecell{LSUN \\ Bedroom \\ 64$\times$64} & \makecell{LSUN \\ Bedroom \\ 256$\times$256} &
                \makecell{CelebA-HQ \\ 1024$\times$1024} \\
                \midrule
                \makecell{DCGAN~\cite{dcgan} \\ + TTUR~\cite{fid}} & 12.5 & - & 57.5 & - \\
                \specialrule{0.1pt}{2pt}{2pt}
                \makecell{WGAN-GP~\cite{WGAN-GP} \\ + TTUR~\cite{fid}} & - & - & 9.5 & - & - \\
                \specialrule{0.1pt}{2pt}{2pt}
                \makecell{IntroVAE~\cite{intro-vae}} & - & - & - & 8.84 & - \\
                \specialrule{0.1pt}{2pt}{2pt}
                PGGAN~\cite{PGGAN} & - & 7.30 & - & 8.34 & \textbf{\underline{7.48}} \\
                \midrule\midrule
                \makecell{Proj. $D$~\cite{projection-discriminator} \\ (our backbone)} & - & 19.55 & - & - & - \\
                \specialrule{0.1pt}{2pt}{2pt}
                \makecell{Ours \\ (N2,M2,S32)} & \textbf{\underline{4.00}} & \textbf{\underline{5.74}} & \textbf{\underline{5.20}} & \textbf{\underline{5.99}}$^*$ & 9.49$^*$ \\
                \bottomrule
            \end{tabular}
            \caption{The FID score suggests that \modelName is competitive with other state-of-the-art generative models. FID scores are measured between 50,000 real and generated samples based on the original implementation provided at {\color{blue}\urlstyle{same}\url{https://github.com/bioinf-jku/TTUR}}. Note that all the FID scores (except proj. $D$) are officially reported numbers. The real samples for evaluation are held-out from training.}
            \label{table:fid-score}
            \vspace{-1em}
        \end{table}
    
    \subsection{Latent Space Continuity} 
    
        To demonstrate the space continuity more precisely, we perform the interpolation experiment in two directions: ``full-images interpolation'' and ``coordinates interpolation''. \blfootnote{\vspace{-1em} We describe the model details in Appendix~\ref{appendix:model-architecture-detail}.}
        
\begin{figure}[t]
            \centering
            \includegraphics[width=\linewidth]{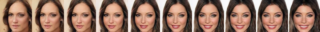} \\
            \includegraphics[width=\linewidth]{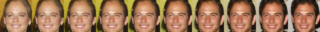} \\
            \includegraphics[width=\linewidth]{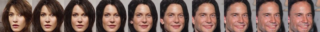} \\
            \caption{The results of full-images interpolation between two latent vectors show that all micro patches are changed synchronously in response to the change of the latent vector. More interpolation results are available in Appendix~\ref{appendix:more-interp}.} %
            \label{fig:interp}
        \end{figure}

        \paragraph{Full-Images Interpolation.} Intuitively, the inter-full-image interpolation is challenging for \modelNamePunc, since all micro patches generated with different spatial coordinates must all change synchronously to make the full-image interpolation smooth. Nonetheless, as shown in Figure~\ref{fig:interp}, we empirically find \modelName can interpolate smoothly and synchronously without producing unnatural artifacts. We randomly sample two latent vectors $z_1$ and $z_2$. With any given interpolation point $z'$ in the slerp-path~\cite{slerp-interpolation} between $z_1$ and $z_2$, the generator uses the full spatial coordinate sequence $\bm{C''}_{\textbf{\textit{full}}}$ to generate all corresponding patches. Then we assemble all the generated micro patches together and form a generated full image $s$.
        
\begin{figure}[t]
            \centering
            \begin{tabular}{cc}
                \raisebox{-.3\height}{\includegraphics[width=0.65\linewidth]{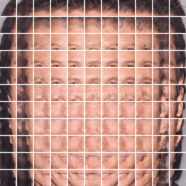}} & \includegraphics[width=0.25\linewidth]{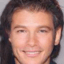} 
            \end{tabular}
            \caption{An example of spatial coordinates interpolation showing the spatial continuity of the micro patches. The spatial coordinates are interpolated between range $[-1, 1]$ of the micro coordinate with a fixed latent vector. More examples are shown in Appendix~\ref{appendix:interp-spatial}.}
            \label{fig:interp-spatial}
            \vspace{-1em}
        \end{figure}

        \paragraph{Coordinates Interpolation.}~\label{exp:interp-spatial} Another dimension of the interpolation experiment is inter-class (\eg between spatial coordinate condition) interpolation with a fixed latent vector. We linearly-interpolate spatial coordinates between $[-1, 1]$ with a fixed latent vector $z$. The results in Figure~\ref{fig:interp-spatial} show that, although we only uniformly sample spatial coordinates within a discrete spatial coordinate set, the spatial coordinates interpolation is still overall continuous. 
        
        An interesting observation is about the interpolation at the position between the eyebrows. In Figure~\ref{fig:interp-spatial}, \modelName does not know the existence of the glabella between two eyes due to the discrete and sparse spatial coordinates sampling strategy. Instead, it learns to directly deform the shape of the eye to switch from one eye to another. This phenomenon raises an interesting discussion, even though the model learns to produce high-quality face images, it still may learn wrong relationships of objects behind the scene.
        
    \subsection{Beyond-Boundary Generation}
    
        \modelName enables a new type of image generation that has never been achieved by GANs before: generate full images that are larger than \textbf{\emph{any}} training sample \textbf{\emph{from scratch}}. In this context, all the generated images are \textbf{\emph{guaranteed}} to be novel and original, since these generated images do not even exist in the training distribution. A supportive evidence is that the generated images have higher resolution than any sample in the training data. In comparison, existing GANs mostly have their output shape fixed after its creation and prove the generator can produce novel samples instead of memorizing real data via interpolating between generated samples.
         \blfootnote{$^*$ The model is not fully converged due to computational resource constraints. One can obtain even lower FID with more GPU-days.}
        
        A shared and interesting behavior of learned manifold of GANs is that, in most cases, the generator can still produce plausible samples with latent vectors slightly out of the training distribution, which we called extrapolation. 
        We empirically observe that with a fixed $z$, extrapolation can be done on the coordinate condition \textit{beyond} the training coordinates distribution. However, as the continuity among patches at these positions is not considered during training, the generated images might show a slight discontinuity at the border. As a solution, we apply a straightforward post-training process (described in Appendix~\ref{appendix:beyond-boundary-generation}) for improving the continuity among patches.
        
\begin{figure}[t]
            \includegraphics[width=\linewidth]{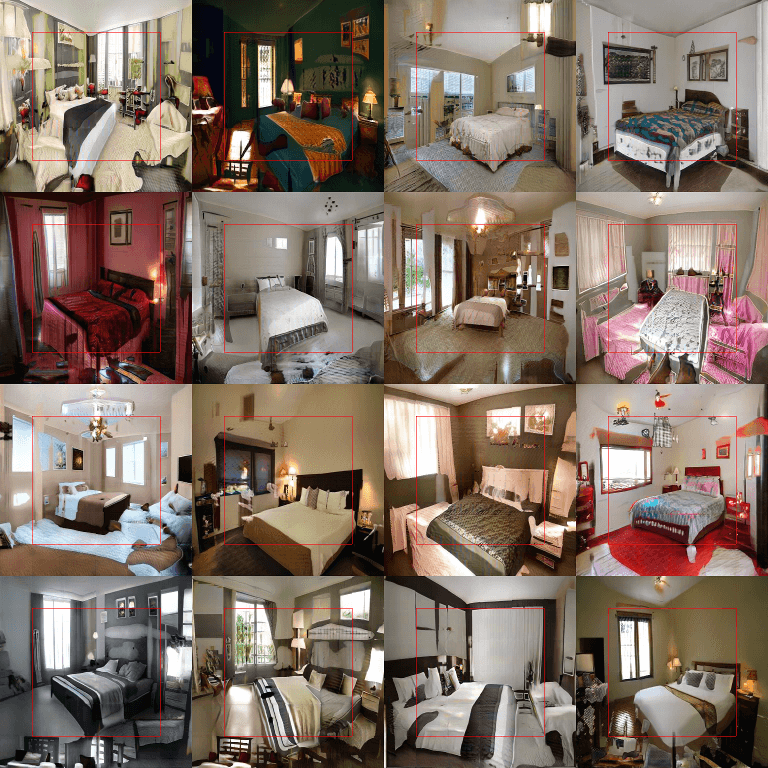}
            \caption{``Beyond-Boundary Generation'' generates additional contents by extrapolating the learned coordinate manifold. Note that the generated samples are $384\times384$ pixels, whereas \textbf{\textit{all}} of the training samples are of a smaller $256\times256$ resolution. The {\color{red} red} box annotates the $256\times256$ region for regular generation without extrapolation.} 
            \label{fig:beyond-boundary-generation}
            \vspace{-1em}
        \end{figure}

        In Figure~\ref{fig:beyond-boundary-generation}, we perform the post-training process on checkpoint of (N4,M4,S64) variant of \modelName that trained on LSUN dataset. Then, we show that \modelName generates high-quality $384\times384$ images: the original size is 256, with each direction being extended by one micro patch (64 pixels), resulting a size of $384\times384$. Note that the model is in fact trained on $256\times256$ images. 
        
    
    \subsection{Panorama Generation \& Partial Generation}
    
        \label{exp:panorama-generation}
        
\begin{figure*}[t]
            \includegraphics[width=\linewidth]{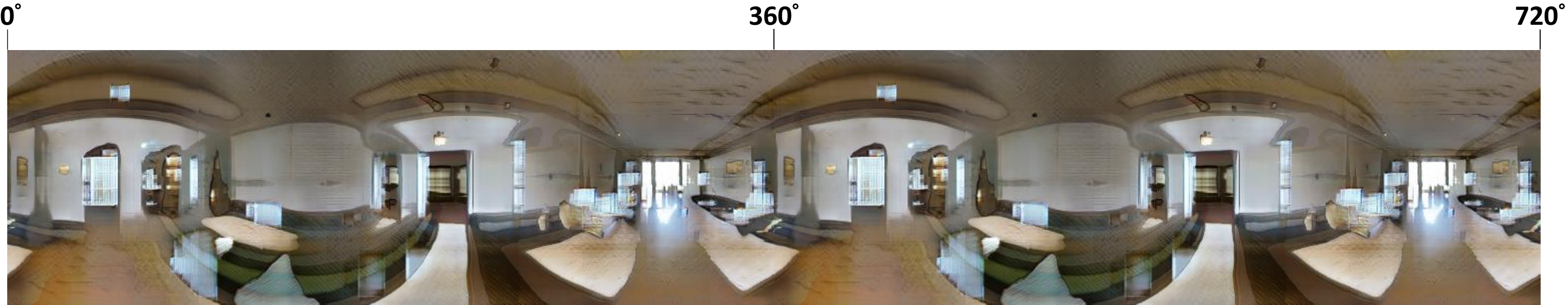}
            \caption{The generated panorama is cyclic in the horizontal direction since \modelName is trained with a cylindrical coordinate system. Here, we paste the same generated panorama twice (from $360^\circ$ to $720^\circ$) to better illustrate the cyclic property of the generated panorama. More generation results are provided in Appendix~\ref{appendix:panorama-samples}.}
            \label{fig:cyclic-panorama}
        \end{figure*}

        Generating panoramas using GANs is an interesting problem but has never been carefully investigated. Different from normal image generation, panoramas are expected to be cylindrical and cyclic in the horizontal direction. However, normal GANs do not have built-in ability to handle such cyclic characteristic if without special types of padding mechanism support~\cite{cube-padding}. In contrast, \modelName is a coordinate-system-aware learning framework. We can easily adopt a cylindrical coordinate system, and generate panoramas that are having ``cyclic topology'' in the horizontal direction as shown in Figure~\ref{fig:cyclic-panorama}. 

        To train \modelName with a panorama dataset under a cylindrical coordinate system, the spatial coordinate sampling strategy needs to be slightly modified. In the horizontal direction, the sampled value within the normalized range $[-1, 1]$ is treated as an angular value $\theta$, and then is projected with $\cos(\theta)$ and $\sin(\theta)$ individually to form a unit-circle on a 2D surface. Along with the original sampling strategy on the vertical axis, a cylindrical coordinate system is formed. 
        
        We conduct our experiment on Matterport3D~\cite{matterport3d} dataset. We first take the sky-box format of the dataset, which consists of six faces of a 3D cube. We preprocess and project the sky-box to a cylinder using Mercator projection, then resize to $768\times 512$ resolution. Since the Mercator projection creates extreme sparsity near the northern and southern poles, which lacks information, we directly remove the upper and lower $1/4$ areas. Eventually, the size of panorama we use for training is $768\times 256$ pixels.

        We also find \modelName has an interesting connection with virtual reality (VR). VR is known to have a tight computational budget due to high frame-rate requirement and high-resolution demand. It is hard to generate full-scene for VR in real time using standard generative models. Some recent VR studies on omnidirectional view rendering and streaming~\cite{vr-streaming-1,vr-streaming-2,vr-streaming-3} focus on reducing computational cost or network bandwidth by adapting to the user's viewport. \modelNamePunc, with the generation-by-parts feature, can easily inherit the same strategy and achieve computation on-demand with respect to the user's viewpoint. Such a strategy can largely reduce unnecessary computational cost outside the region of interest, thus making image generation in VR more applicable.
        
    \subsection{Patch-Guided Image Generation}
        
        We further explore an interesting application of \modelName named ``Patch-Guided Image Generation''. By training an extra auxiliary network $Q$ within $D$ that predicts the latent vector of each generated macro patch $s'$, the discriminator is able to find a latent vector $z_{\textit{est}} = Q(x')$ that generates a macro patch similar to a provided real macro patch $x'$. Moreover, the estimated latent vector $z_{\textit{est}}$ can be applied to the full-image generation process, and eventually generates an image that is partially similar to the original real macro patch, while globally coherent. 
        
\begin{figure}[t]
            \centering
            \parbox[b]{.075\linewidth}{\rotatebox[origin=l]{90}{\hspace{0.4em} Macro Patch}}%
            \includegraphics[width=0.27\linewidth]{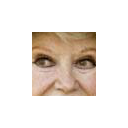}%
            \hspace{0.2em}%
            \includegraphics[width=0.27\linewidth]{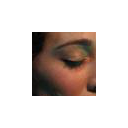}%
            \hspace{0.2em}%
            \includegraphics[width=0.27\linewidth]{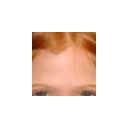}
            
            \vspace{0.1em}
            
            \parbox[b]{.075\linewidth}{\rotatebox[origin=l]{90}{\hspace{0.2em} Partial Conv.}}%
            \includegraphics[width=0.27\linewidth]{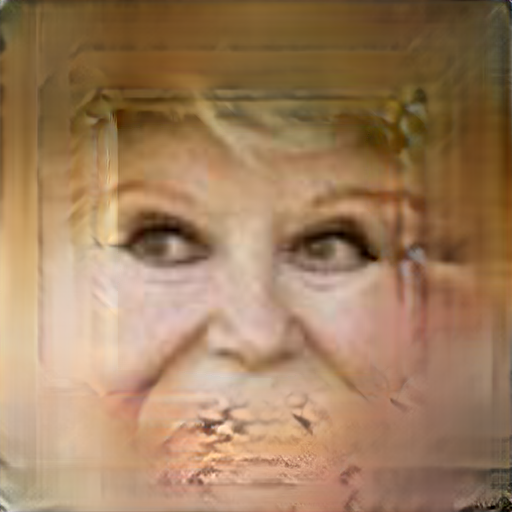}%
            \hspace{0.2em}%
            \includegraphics[width=0.27\linewidth]{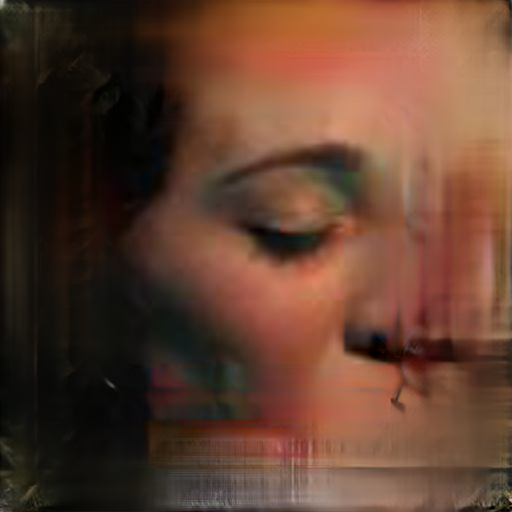}%
            \hspace{0.2em}%
            \includegraphics[width=0.27\linewidth]{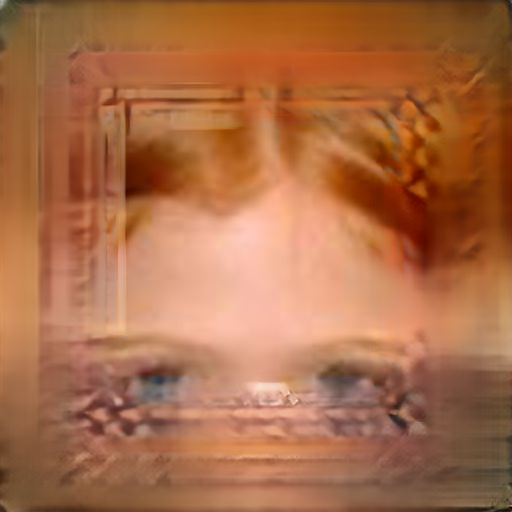}
            
            \vspace{0.1em}
            
            \parbox[b]{.075\linewidth}{\rotatebox[origin=l]{90}{\hspace{1.5em} Ours}}%
            \includegraphics[width=0.27\linewidth]{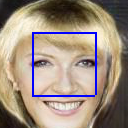}%
            \hspace{0.2em}%
            \includegraphics[width=0.27\linewidth]{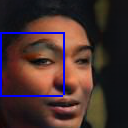}%
            \hspace{0.2em}%
            \includegraphics[width=0.27\linewidth]{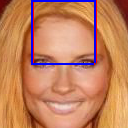}
            
            \caption{Patch-guided image generation loosely retains the local structures from the original image and make the full image still globally coherent. The quality outperforms the partial convolution~\cite{partial-conv}. The {\color{blue} blue} boxes visualize the predicted spatial coordinates $A(x')$, while the {\color{red} red} boxes indicate the ground truth coordinates $c'$. Note that the generated images are \textbf{not} expected to be identical to the original real images. More examples are provided in Appendix~\ref{appendix:patch-guided-image-generation}.}
            \label{fig:patch-guided-image-generation}
         \end{figure}

        This application shares similar context to some bijection methods~\cite{ali,bigan,BEGAN-CS}, despite \modelName estimates the latent vector with a single macro patch instead of the full image. In addition, the application is also similar to image restoration~\cite{partial-conv,inpaintingC,inpaintingB} or image out-painting~\cite{out-painting}. However, these related applications heavily rely on the information from the surrounding environment, which is not fully accessible from a single macro patch. In Figure~\ref{fig:patch-guided-image-generation}, we show that our method is robust to extremely damaged images. More samples and analyses are described in Appendix~\ref{appendix:patch-guided-image-generation}.
        
    \subsection{Computation-Friendly Generation}
    
        Recent studies in high-resolution image generation~\cite{PGGAN, R1-regularizer,intro-vae} have gained lots of success; however, a shared conundrum among these existing approaches is the computation being memory hungry. Therefore, these approaches make some compromises to reduce memory usage~\cite{PGGAN,R1-regularizer}. Moreover, this memory bottleneck cannot be easily resolved without specific hardware support, which makes the generation of \textit{over} $1024\times 1024$ resolution images difficult to achieve. These types of high-resolution images are commonly seen in panoramas, street views, and medical images. 
    
\begin{figure}[h]
            \centering
            \includegraphics[width=\linewidth]{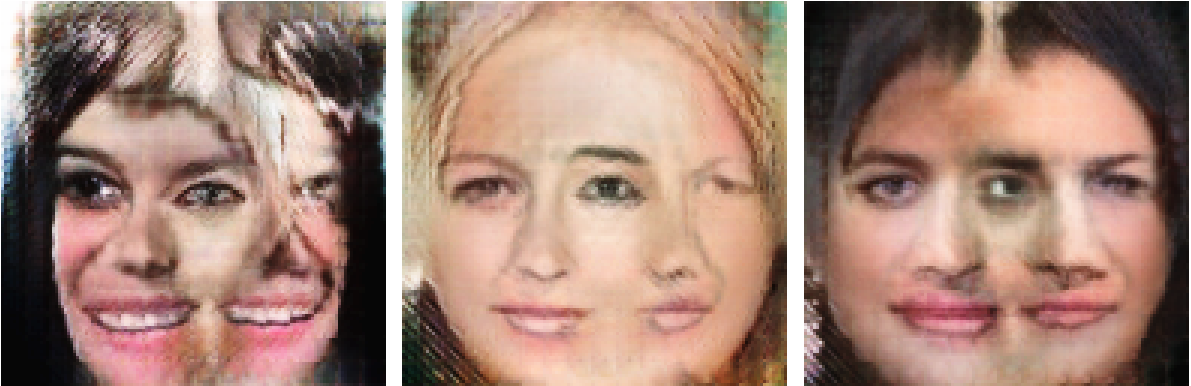}
            \caption{Examples to show that with macro patches smaller than 1/16 of the full image causes \modelName to learn incorrect spatial relation. Note that this value may vary due to the nature (local structure, texture, etc) of each dataset being different.}
            \label{fig:macro-too-small}
        \end{figure}

        In contrast, \modelName only requires partial views of the full image for both training and inference. Note that the memory consumption for training (and making inference) GANs grows approximately linearly with respect to the image size. Due to using only partial views, \modelName changes the growth in memory consumption to be associated with the size of a macro patch, not the full image. For instance, on the CelebA $128\times128$ dataset, the (N2,M2,S16) setup of \modelName reduces memory requirement from 17,184 MB (our projection discriminator backbone) to 8,992 MB (\ie, 47.7\% reduction), with a batch size 128. However, if the size of a macro patch is too small, \modelName will be misled to learn incorrect spatial relation; in Figure~\ref{fig:macro-too-small}, we show an experiment with a macro patch of size $32\times32$ and a micro patch size of $16\times16$. Notice the low quality (\ie, duplicated faces). Empirically, the minimum requirement of macro patch size varies for different datasets; for instance, \modelName does not show similar poor quality in panorama generation in Section~\ref{exp:panorama-generation}, where the macro patch size is 1/48 of the full panorama. Future research on a) how to mitigate such effects (for instance, increase the receptive field of $D$ without harming performance) and b) how to evaluate a proper macro patch size, may further advance the generation-by-parts property particularly in generating large field-of-view data.
        
    \subsection{Ablation Study}
        \label{exp:ablation-study}
        
        \begin{table}
            \small
            \centering
            \begin{tabular}{l c}
                \toprule \\ [-1.2em]
                \textbf{Model} & \textbf{best FID (150 epochs)} \\ [0.1em]
                \hline
                \modelName (cont. sampling)      & 6.13 \\
                \modelName + optimal $D$         & 4.05 \\
                \modelName + optimal $G$         & 6.12 \\
                Multiple $G$                     & 7.26 \\
                \hline \\ [-0.9em]
                \modelName (N2,M2,S16)           & 4.87 \\
                \bottomrule
            \end{tabular}
            \caption{The ablation study shows that \modelName (N2,M2,S16) can converge well with little trade-off in convergence speed on CelebA $64\times64$ dataset.}
            \label{table:ablation}
            \vspace{-1em}
        \end{table}
        
        In Table~\ref{table:ablation}, the ablation study aims to analyze the \textbf{trade-offs} of each component of \modelNamePunc. We perform experiments in CelebA $64\times 64$ with four ablation configurations: ``continuous sampling'' demonstrates that using continuous uniform sampling strategy for spatial coordinates during training will result in moderate generation quality drop; ``optimal $D$'' lets the discriminator directly discriminate the full image while the generator still generates micro patches; ``optimal $G$'' lets the generator directly generate the full image while the discriminator still discriminates macro patches; ``multiple $G$'' trains an individual generator for each spatial coordinate. 
        
        We observe that, surprisingly, despite the convergence speed is different, ``optimal discriminator'', \modelNamePunc, and ``optimal generator'' (ordered by convergence speed from fast to slow) can all achieve similar FID scores if with sufficient training time. The difference in convergence speed is expected since ``optimal discriminator'' provides the generator with more accurate and global adversarial loss. In contrast, the ``optimal generator`` has relatively more parameters and layers to optimize, which causes the convergence speed slower than \modelNamePunc. Lastly, the ``multiple generators'' setting cannot converge well. Although it can also concatenate micro patches without obvious seams as \modelName does, the full-image results often cannot agree and are not coherent. More experimental details and generated samples are shown in Appendix~\ref{appendix:ablation}.
        
    \subsection{Non-Aligned Dataset}
    
        \begin{figure}[t]
            \centering
            \adjustbox{trim=0 0 0 0,clip}%
            {\includegraphics[width=\linewidth]{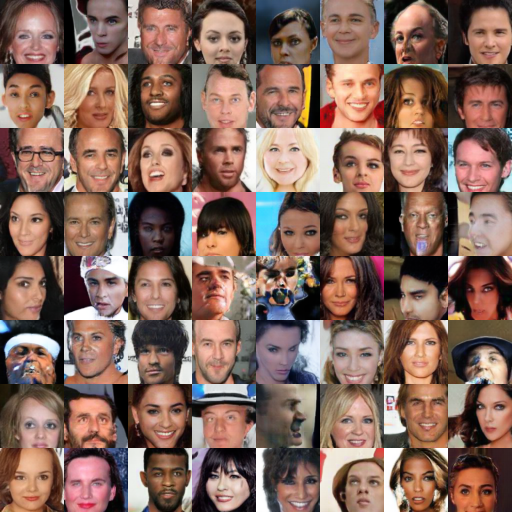}}
            \vspace{-1.5em}
            \caption{COCO-GAN can learn and synthesis samples with diverse position on the non-aligned \textit{Celeba-syn}.}
            \label{fig:celeba-syn}
            \vspace{-1.5em}
        \end{figure}
    
        It is easy to get confused that the coordinate system would restrain COCO-GAN from learning on less aligned datasets. In fact, this is completely not true. For instance, the bedroom category of LSUN, the location, size and orientation of the bed are very dynamic and non-aligned. On the other hand, the Matterport3D panoramas are completely non-aligned in the horizontal direction. 
        
        To further resolve all the potential concerns, we propose \textit{CelebA-syn}, which applies a random displacement on the raw data (different from data augmentation, this preprocessing directly affects the dataset) to mess up the face alignment. We first trim the raw images to 128$\times$128. The position of the upper-left corner is sampled by $(x,y) = (25+dx, 50+dy)$, where $dx\sim\mathcal{U}(-25,25)$ and $dy\sim\mathcal{U}(-25,25)$. Then we resize the trimmed images to 64$\times$64 for training. As shown in Figure~\ref{fig:celeba-syn}, COCO-GAN can stably create reasonable samples of high diversity (also notice the high diversity at the eye positions).
        
\section{Related Work}

    Generative Adversarial Network (GAN)~\cite{GANs} and its conditional variant~\cite{cGAN} have shown their potential and flexibility to many different tasks. Recent studies on GANs are focusing on generating high-resolution and high-quality synthetic images in different settings. For instance, generating images with $1024\times 1024$ resolution ~\cite{PGGAN,R1-regularizer}, generating images with low-quality synthetic images as condition~\cite{apple-eye}, and by applying segmentation maps as conditions~\cite{pix2pix-HD}. However, these prior works share similar assumptions: the model must process and generate the full image in a single shot. This assumption consumes an unavoidable and significant amount of memory when the size of the targeting image is relatively large, and therefore makes it difficult to satisfy memory requirements for both training and inference. Searching for a solution to this problem is one of the initial motivations of this work.

    \modelName shares some similarities to Pixel-RNN~\cite{pixel-rnn}, which is a pixel-level generation framework while \modelName is a patch-level generation framework. Pixel-RNN transforms the image generation task into a sequence generation task and maximizes the log-likelihood directly. In contrast, \modelName aims at decomposing the computation dependencies between micro patches across the spatial dimensions, and then uses the adversarial loss to ensure smoothness between adjacent micro patches.
    
    CoordConv~\cite{coord-conv} is another similar method but with fundamental differences. CoordConv provides spatial positioning information directly to the convolutional kernels in order to solve the coordinate transform problem and shows multiple improvements in different tasks. In contrast, \modelName uses spatial coordinates as an auxiliary task for the GANs training, which enforces both the generator and the discriminator to learn coordinating and correlations between the generated micro patches. We have also considered incorporating CoordConv into \modelNamePunc. However, empirical results show little visual improvement.
    
    

\section{Conclusion and Discussion}

    In this paper, we propose \modelNamePunc, a novel GAN incorporating the conditional coordination mechanism. \modelName enables ``generation by parts'' and demonstrates the generation quality being competitive to state-of-the-arts. \modelName also enables several new applications such as ``Beyond-Boundary Generation'' and ``Panorama Generation'', which serve as intriguing directions for future research on leveraging the learned coordinate manifold for (a) tackling with large field-of-view generation and (b) reducing computational requisition.
    
    
    
    Particularly, given a random latent vector, Beyond-Boundary Generation generates images larger than \textit{any} training sample by extrapolating the learned coordinate manifold, which is enabled exclusively by \modelNamePunc. Future research on extending this property to other tasks or applications may further take advantage of such an out-of-distribution generation paradigm.
    
    We show that \modelName produces $128\times128$ images with micro patches as small as $4\times4$ pixels. The overall FID score slightly degrades due to the small micro patch size. Further studies on the relationship between the patch size and generation stability are left as a straight-line future work.
    
    Although \modelName has achieved a high generation quality comparable to state-of-the-art GANs, for several generated samples we still observe that the local structures may be discontinued or mottled. This suggests further studies on additional refinements or blending approaches that could be applied on \modelName for generating more stable and reliable samples.
    
    

\section{Acknowledgement} 

    We sincerely thank David Berthelot and Mong-li Shih for the insightful suggestions and advice. We are grateful to the National Center for High-performance Computing for computer time and facilities. Hwann-Tzong Chen was supported in part by MOST grants 107-2634-F-001-002 and 107-2218-E-007-047.

\onecolumn
\icmltitle{\vspace{-0.5em} (Appendix) COCO-GAN: Generation by Parts via Conditional Coordinating}
\begin{appendices}

\section{\modelName during Testing Phase}
    \label{appendix:model-testing-phase}
    \begin{figure}[H]
        \centering
        \includegraphics[width=\linewidth]{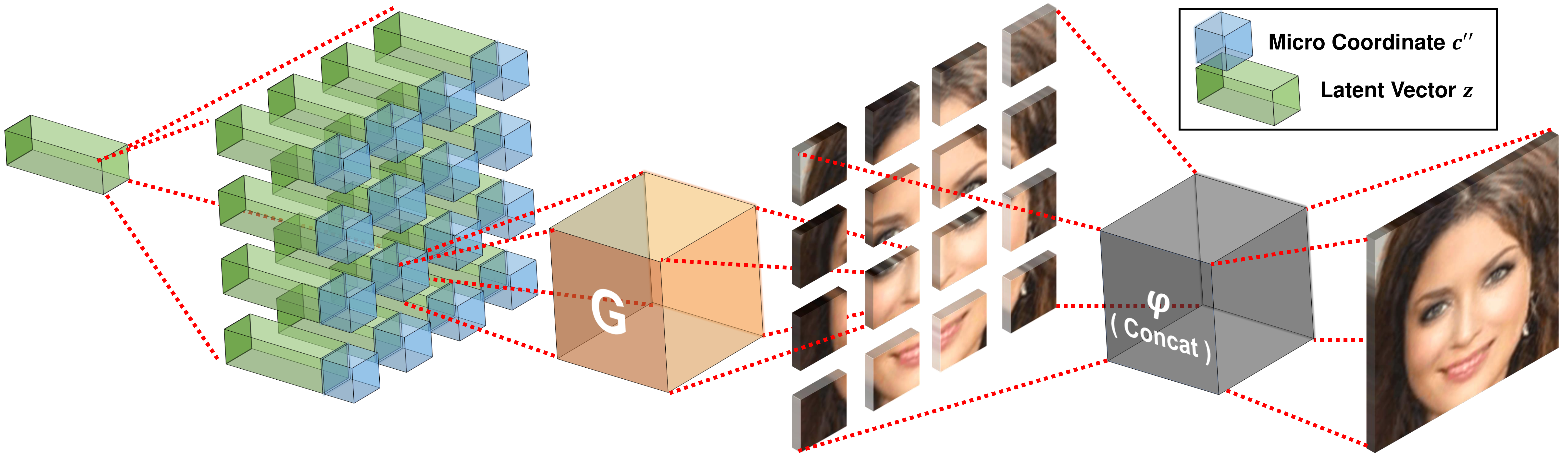}
        \caption{An overview of \modelName during testing phase. The micro patches generated by $G$ are directly combined into a full image as the final output.}
        \label{fig:model-testing-phase}
        \vspace{-1em}
    \end{figure}
    
\section{Symbols}
    \label{appendix:symbols}
    \begin{table}[h]
        \vspace{-1em}
        \centering
        \small
        \setlength\tabcolsep{3pt}
        \begin{tabular}{c | c | l | l | l}
            \toprule \\ [-1.2em]
            \textbf{Group} & \textbf{Symbol} &  \textbf{Name} & \textbf{Description} & \textbf{Usage}\\
            \specialrule{0.1pt}{2pt}{2pt}
            \multirow{4}{*}{Model} & $G$ & Generator & Generates micro patches.& $s''=G(z,c'')$ \\
             & $D$ & Discriminator & Discriminates macro patches. & $D(\varphi(G(z,\bm{C''})))$ \\
             & $A$ & Spatial prediction head & Predicts coordinate of a given macro patch. & $\hat{c}'=A(x')$ \\
             & $Q$ & $^\dagger$Content prediction head & Predicts latent vector of a given macro patch. & $z_{est}=Q(s')$ \\
            \specialrule{0.1pt}{2pt}{2pt}
            \multirow{2}{*}{\makecell{Heuristic \\ Function}} &  $\varphi$ & Merging function & Merges multiple $s''$ to form a $s'$ or $s$. & $s'=\varphi(G(z,\bm{C''}))$ \\
             & $\psi$ & Cropping function & Crops $x'$ from $x$. Corresponding to $\varphi$. & $x'=\psi(x, c')$ \\
            \specialrule{0.1pt}{2pt}{2pt}
            \multirow{6}{*}{Variable} & $z$ & Latent vector & Latent variable shared among $s''$ generation. & $s''=G(z, c'')$ \\
             & $z_{est}$ & $^\dagger$Predicted $z$ & Predicted $z$ of a given macro patch. & $L_{Q}=\expectation\left[\,\|z-z_{est}\|_1\,\right]$ \\
             & $c'$ & Macro coordinate & Coordinate for macro patches on $D$ side. & $L_{S}=\expectation\left[\,\|c'-\hat{c}'\|_2\,\right]$ \\
             & $c''$ & Micro coordinate & Coordinate for micro patches on $G$ side. & $s''=G(z,c'')$ \\
             & $\hat{c}'$ & Predicted $c'$ & Coordinate predicted by $A$ with a given $x'$. & $L_{S}=\expectation\left[\,\|c'-\hat{c}'\|_2\,\right]$ \\
             & $\bm{C''}$ & Matrix of $c''$ & The matrix of $c''$ used to generate $\bm{S''}$. & $s' = \varphi(G(z,\bm{C''}))$ \\
            \specialrule{0.1pt}{2pt}{2pt}
            \multirow{6}{*}{Data} & $x$ & Real full image & Full resolution data, never directly used. & $x'=\psi(x, c')$ \\
             & $x'$ & Real macro patch & A macro patch of $x$ which $D$ trains on. & $\text{adv}_{x'}=D(\psi(x, c'))$ \\
             & $s'$ & Generated macro patch & Composed by $s''$ generated with $\bm{C''}$. & $\text{adv}_{s'}=D(s')$ \\
             & $s''$ & Generated micro patch & Smallest data unit generated by $G$. & $s''=G(z,c'')$ \\
             & $\bm{S}''$ & Matrix of $s''$ & Matrix of $s''$ generated by $\bm{C''}$. & $\bm{S''}=G(z,\bm{C''})$ \\
             & $\hat{s}'$ & Interpolated macro patch & Interpolation between random $x'$ and $s'$. & $\hat{s}' = \epsilon \, s' + (1-\epsilon) \, x'$, which $\epsilon\sim\left[0,1\right]$ \\
            \specialrule{0.1pt}{2pt}{2pt}
            \multirow{4}{*}{Loss} & $L_{W}$ & WGAN loss & The patch-level WGAN loss. &  $L_{W}=\expectation \left[D(x')\right] - \expectation \left[D(s')\right]$ \\
             & $L_{GP}$ & Gradient penalty loss & The gradient penalty loss to stabilize training. & $L_{GP} = \expectation \left[ ( \| \nabla_{\hat{s}'} D(\hat{s}') \|_2 - 1 )^2 \right]$ \\
             & $L_{S}$ & Spatial consistency loss & Consistency loss of coordinates. & $L_S = \expectation \left[ \| c' - A(x') \|_2 \right]$\\
             & $L_{Q}$ & $^\dagger$Content consistency loss & Consistency loss of latent vectors. &  $L_{Q} = \expectation \left[ \| z - Q(s') \|_1 \right]$\\
            \specialrule{0.1pt}{2pt}{2pt}
            \multirow{2}{*}{\makecell{Hyper-\\parameter}} & $\alpha$ & Weight of $L_{S}$ & Controls the strength of $L_{S}$ (we use 100). & \multirow{2}{*}{\scriptsize$\begin{cases}
                L_W + \lambda L_{GP} + \alpha L_S, & \text{for} \, D, \\
                - L_W + \alpha L_S,        & \text{for} \, G.
            \end{cases}$} \\
             & $\lambda$ & Weight of $L_{GP}$ & Controls the strength of $L_{GP}$ (we use 10). & \\
            \specialrule{0.1pt}{2pt}{2pt}
            \multirow{2}{*}{\makecell{Testing \\ Only}} & $s$ & Generated full image & Composed by $s''$ generated with $\bm{C''}_{Full}$. & $s = \varphi(G(z,\bm{C''}_{Full}))$ \\
             & $\bm{C''}_{Full}$ & Matrix of $c''$ for testing & The matrix of $c''$ used during testing. & $s = \varphi(G(z,\bm{C''}_{Full}))$ \\
            \bottomrule
        \end{tabular}
        \vspace{-1em}
    \end{table}
    {\small
    $^\dagger$ Only used in ``Patch-Guided Image Generation'' application.
    }
    
\clearpage
    
\section{Experiments Setup and Model Architecture Details}

    \label{appendix:model-architecture-detail}
        
\begin{figure}[H]
        \vspace{-1em}
        \hfill%
        \subfloat[Generator Overall Architecture]{\includegraphics[width=0.35\linewidth]{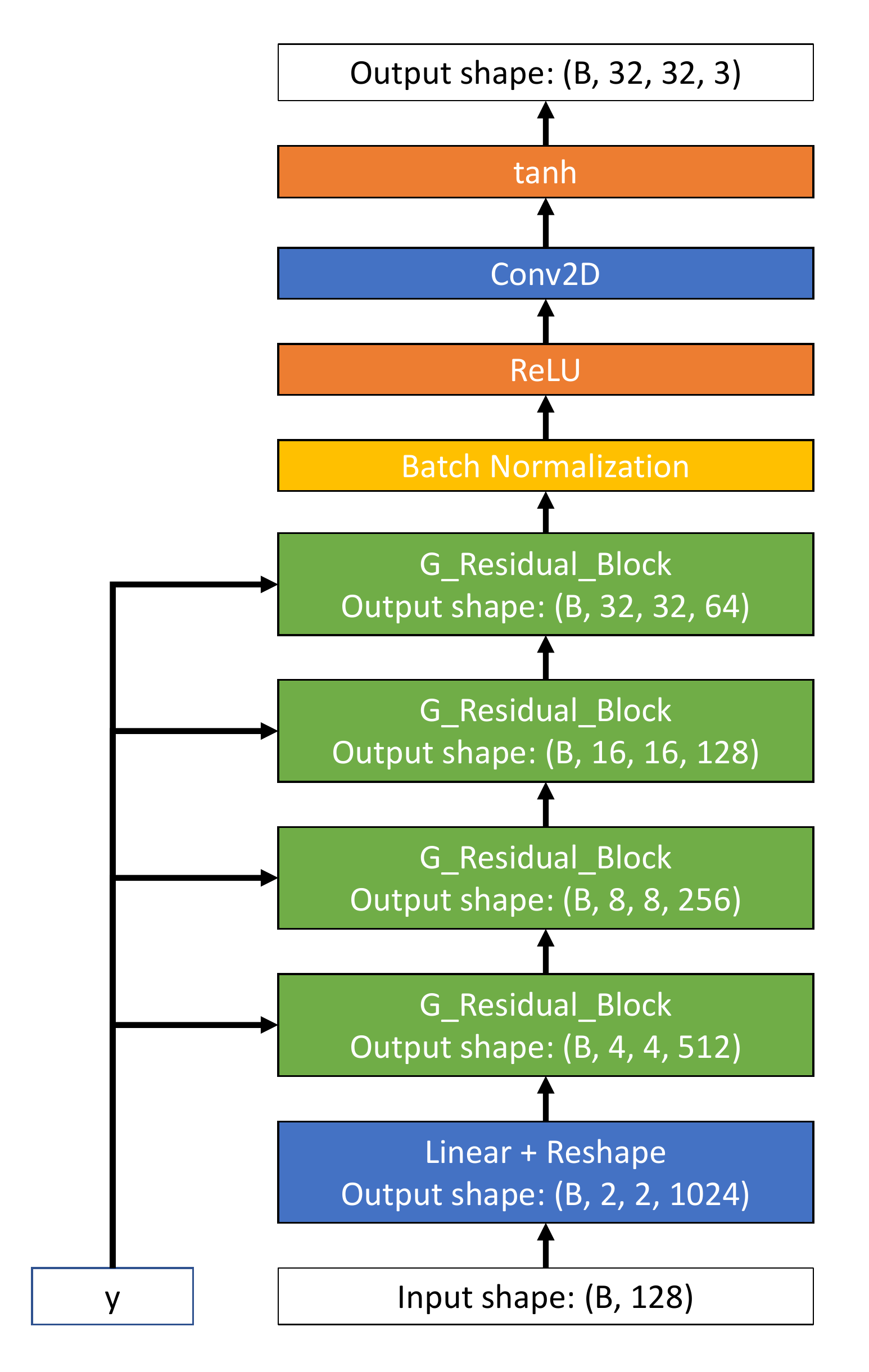}}%
        \hfill%
        \subfloat[Generator Residual Block]{\includegraphics[width=0.56\linewidth]{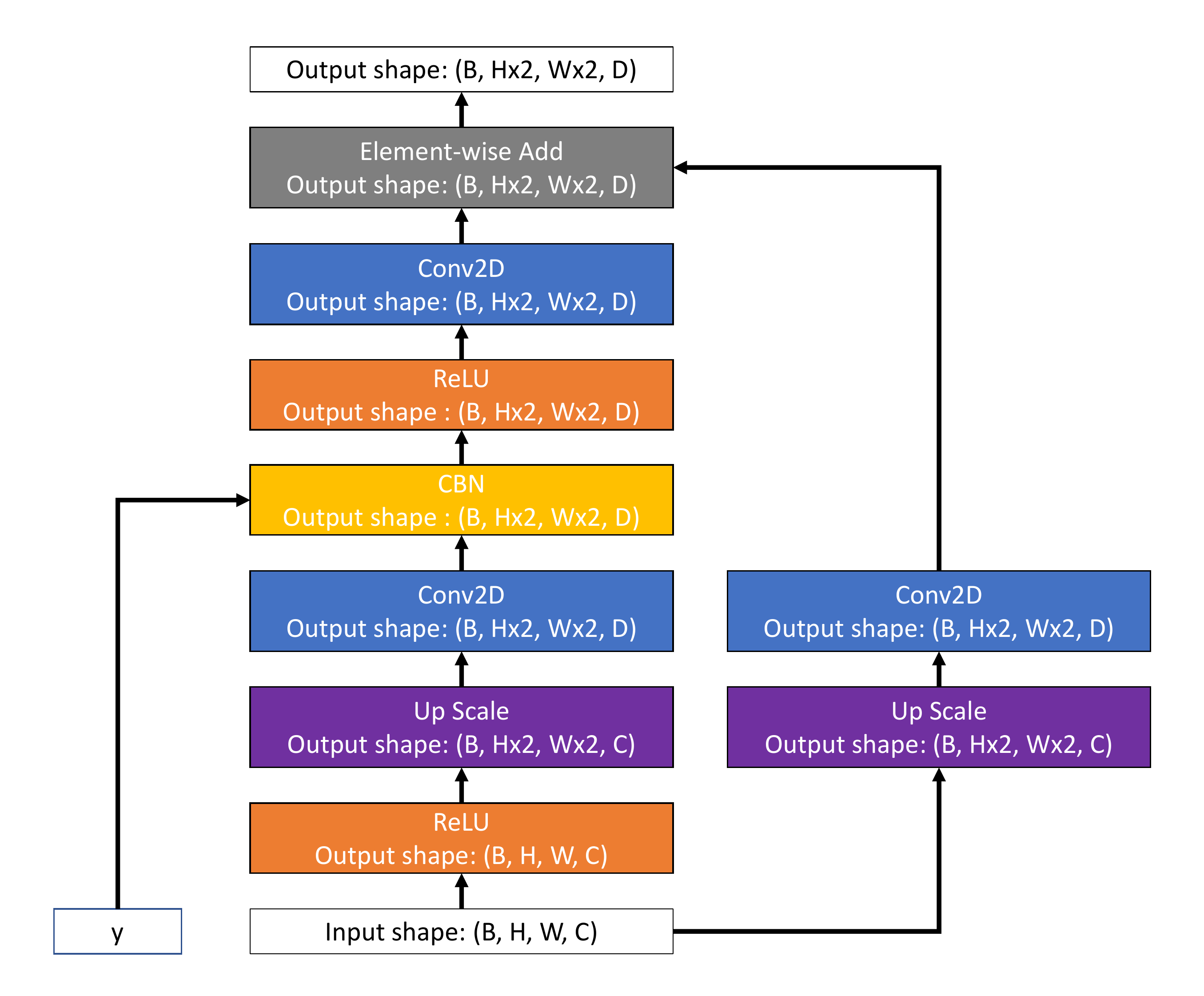}}%
        \hfill
        \caption{The detailed generator architecture of \modelName for generating micro patches with a size of $32\times 32$ pixels.}
        \label{fig:architecture-detail-G}
        \vspace{-1em}
    \end{figure}

\begin{figure}[H]
        \centering
        \subfloat[Discriminator Overall Architecture]{\includegraphics[width=0.56\linewidth]{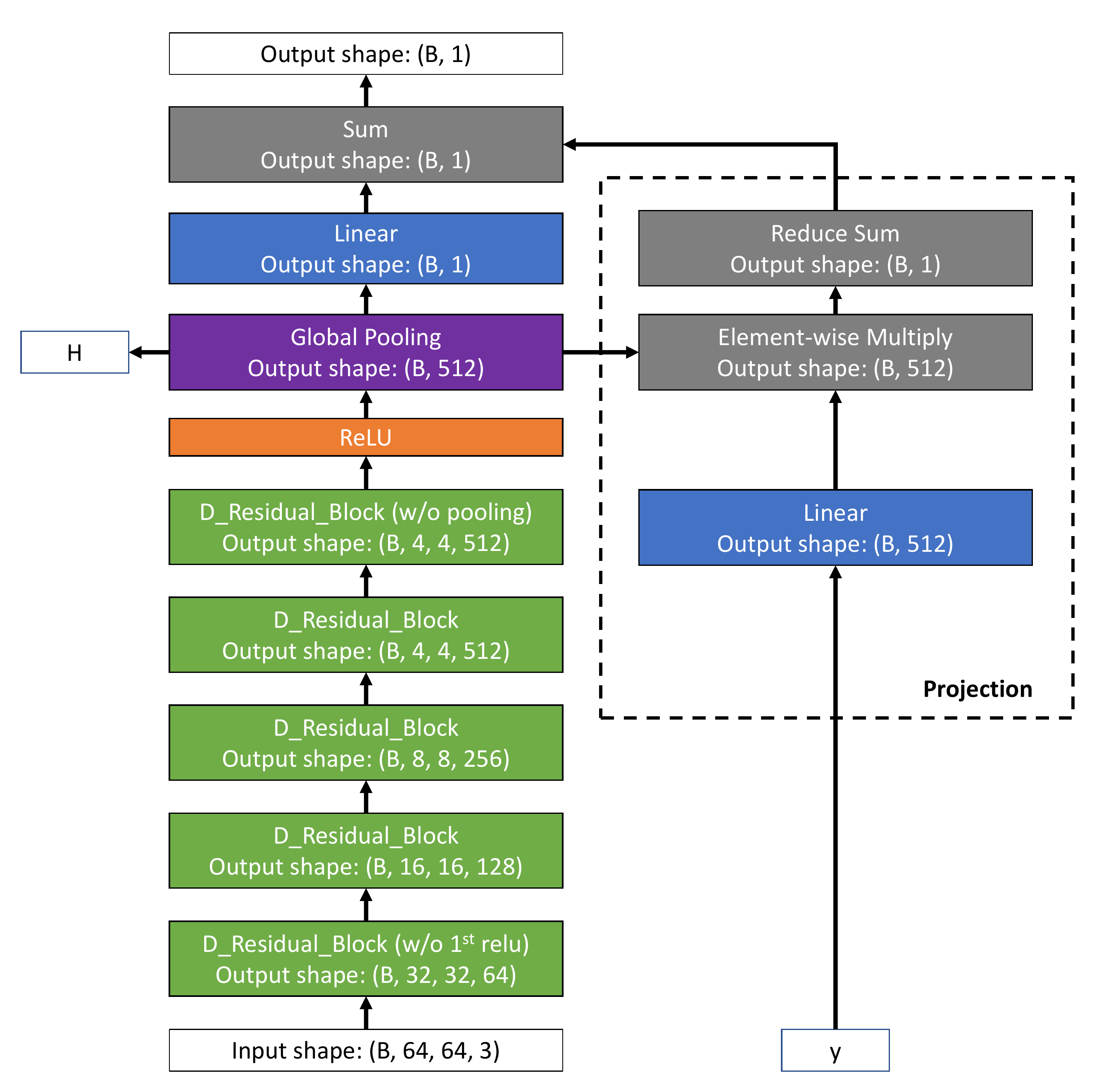}}%
        \hfill%
        \subfloat[Discriminator Residual Block]{\includegraphics[width=0.43\linewidth]{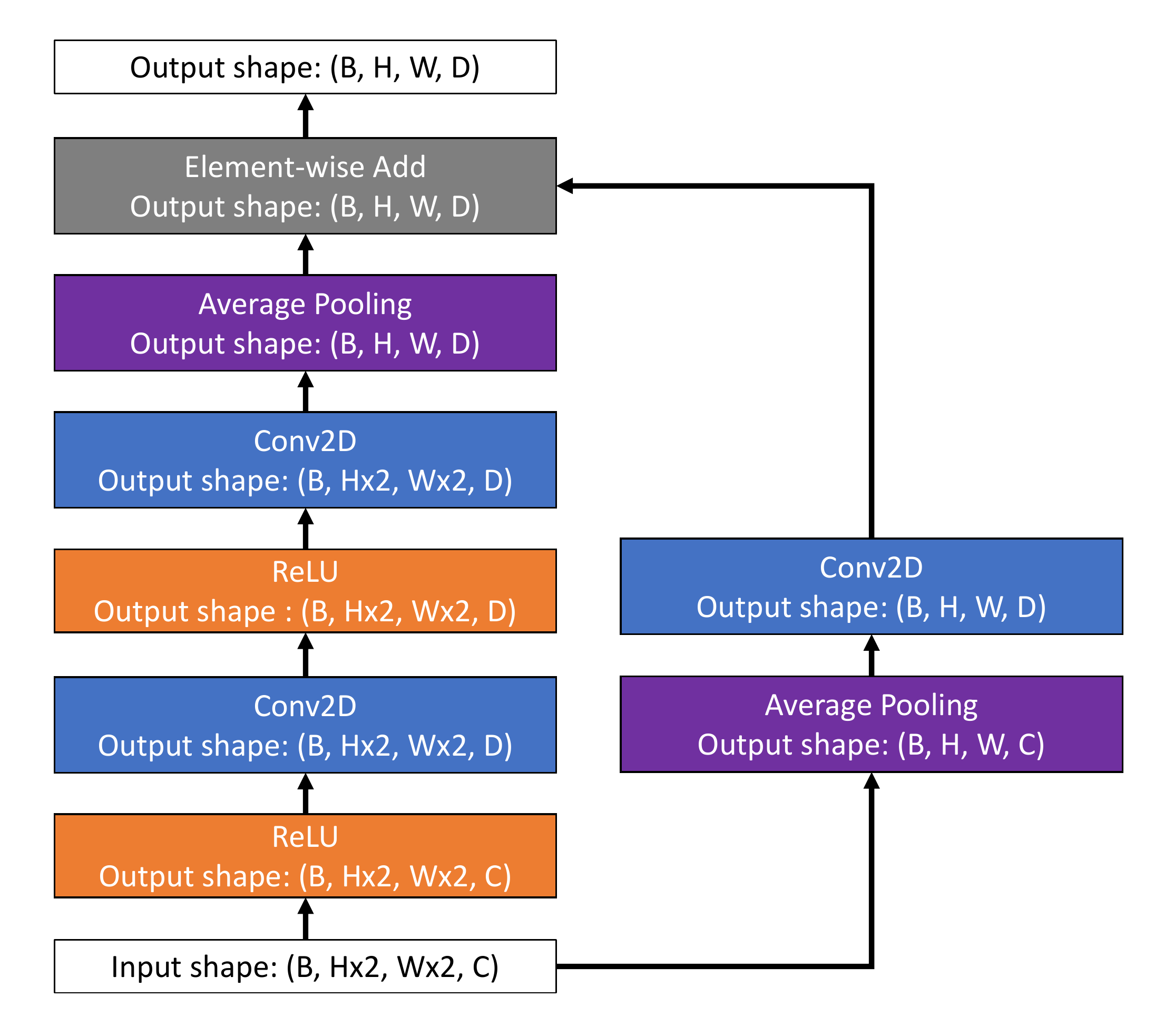}}%
        \vspace{2em}
        \subfloat[Discriminator Auxiliary Head]{
            \includegraphics[width=0.3\linewidth]{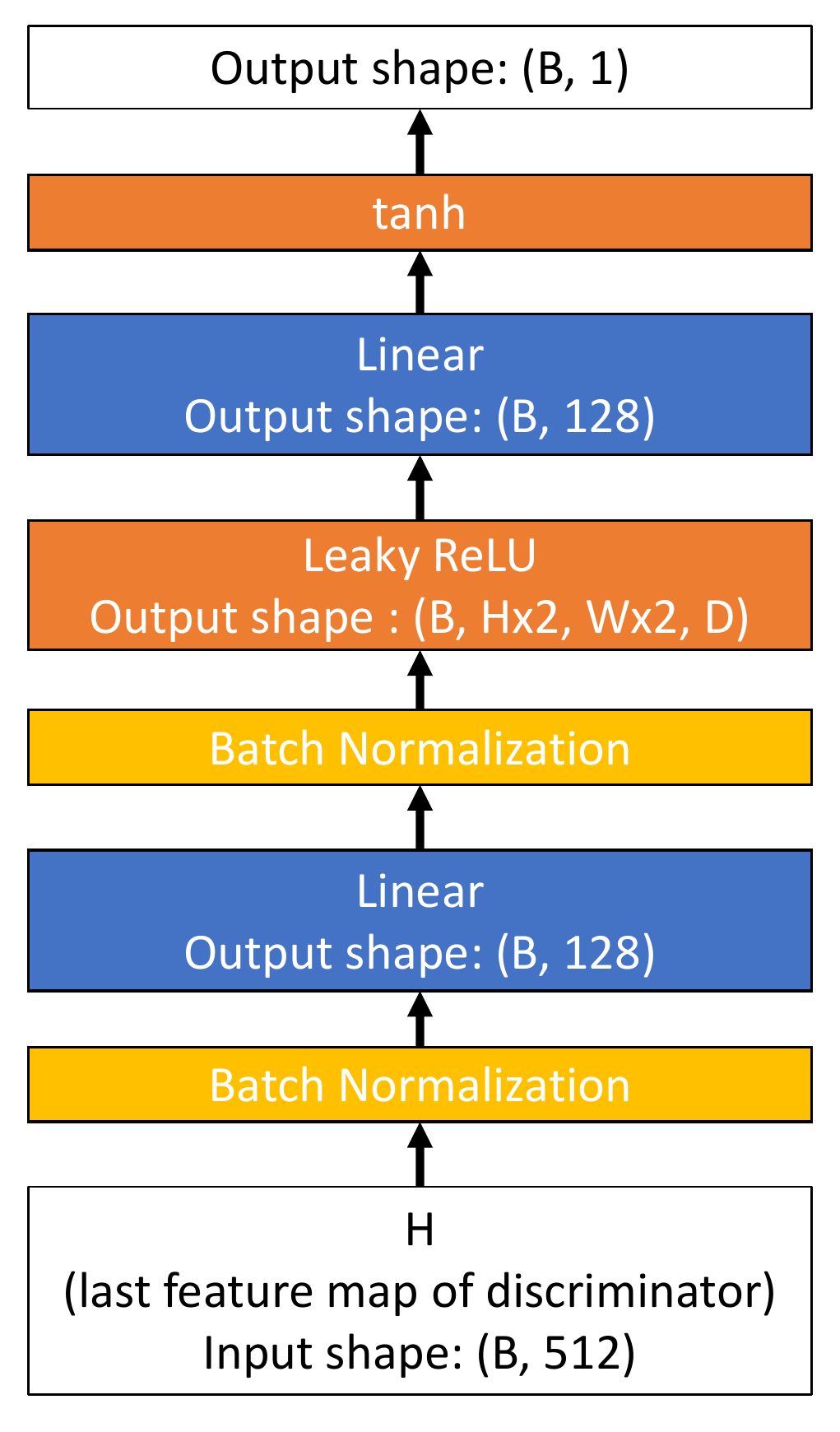}
            \label{fig:aux-head-structure}
        }
        \vspace{1em}
        
        \caption{The detailed discriminator architecture of \modelName for discriminate macro patches with a size of $64\times 64$ pixels. Both the content vector prediction head ($Q$) and the spatial condition prediction head use the same structure shown in (c).}
        \label{fig:architecture-detail-D}
    \end{figure}

    \noindent \textbf{Architecture.} Our $G$ and $D$ design uses projection discriminator~\cite{projection-discriminator} as the backbone and adding class-projection to the discriminator. All convolutional and feed-forward layers of generator and discriminator are added with the spectral-normalization~\cite{SN} as suggested in \cite{SAGAN}. Detailed architecture diagram is illustrated in Figure~\ref{fig:architecture-detail-G} and Figure~\ref{fig:architecture-detail-D}. Specifically, we directly duplicate/remove the last residual block if we need to enlarge/reduce the size of output patch. However, for (N8,M8,S8) and (N16,M16,S4) settings, since the model becomes too shallow, we keep using (N4,M4,S16) architecture, but without strides in the last one and two layer(s), respectively.

    \vspace{0.5em} \noindent \textbf{Conditional Batch Normalization (CBN).} We follow the projection discriminator that employs CBN~\cite{CBN,cont-CBN} in the generator. The concept of CBN is to normalize, then modulate the features by conditionally produce $\gamma$ and $\beta$ that used in conventional batch normalization, which computes $o_K = ( (i_K - \mu_K) / \sigma_K ) * \gamma + \beta$ for the K-th input feature $i_K$, output feature $o_K$, feature mean $\mu_K$ and feature variance $\sigma_K$. However, in the COCO-GAN setup, we provide both spatial coordinate and latent vector as conditional inputs, which both having real values instead of common discrete classes. As a result, we create two MLPs, $\mathrm{MLP}_{\gamma}(z,c)$ and $\mathrm{MLP}_{\beta}(z,c)$, for each CBN layer, that conditionally produces $\gamma$ and $\beta$.
    
    \vspace{0.5em} \noindent \textbf{Hyperparameters.} For all the experiments, we set the gradient penalty weight $\lambda=10$ and auxiliary loss weight $\alpha=100$. We use Adam~\cite{Adam} optimizer with $\beta_1 = 0$ and $\beta_2 = 0.999$ for both the generator and the discriminator. The learning rates are based on the Two Time-scale Update Rule (TTUR)~\cite{fid}, setting $0.0001$ for the generator and $0.0004$ for the discriminator as suggested in \cite{SAGAN}. We do not specifically balance the generator and the discriminator by manually setting how many iterations to update the generator once as described in the WGAN paper \cite{WGAN}.
    
    \vspace{0.5em} \noindent \textbf{Coordinate Setup.} For the micro coordinate matrix $\bm{C}''_{(i,j)}$ sampling, although COCO-GAN framework supports real-valued coordinate as input, however, with sampling only the discrete coordinate points that is used in the testing phase will result in better overall visual quality. As a result, all our experiments select to adopt such discrete sampling strategy. We show the quantitative degradation in the ablation study section. To ensure that the latent vectors $z$, macro coordinate conditions $c'$, and micro coordinate conditions $c''$ share the similar scale, which $z$ and $c''$ are concatenated before feeding to $G$. We normalize $c'$ and $c''$ values into range $[-1, 1]$, respectively. For the latent vectors $z$ sampling, we adopts uniform sampling between $[-1, 1]$, which is numerically more compatible with the normalized spatial condition space.
    

\section{Example of Coordinate Design}
    \label{appendix:coordinate-example}
    \begin{figure}[H]
        \centering
        \subfloat[\fontsize{8}{10}\selectfont Implementations used in this paper with (Left) P4x4, (Middle) P8x8 and (Right) P16x16.]{\includegraphics[width=0.59\linewidth]{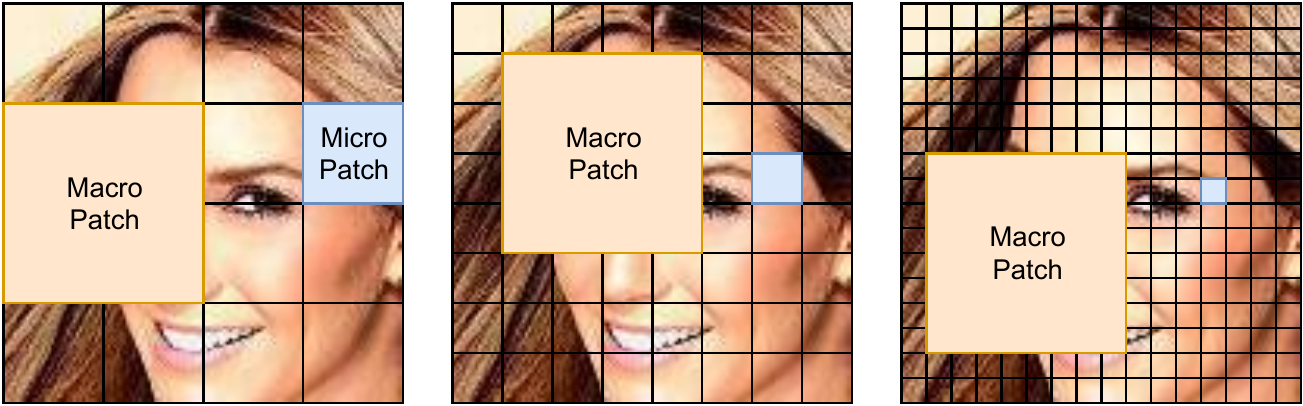}} %
        \hfill%
        \subfloat[\fontsize{8}{10}\selectfont Other possible implementations (not used in this paper).]{\includegraphics[width=0.39\linewidth]{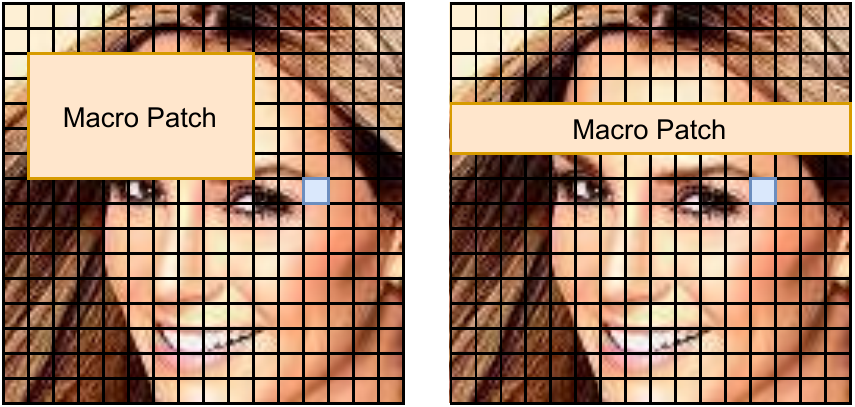}}
        \caption{We showcase some of the coordinate systems: (a) implementations we used in our experiments, and (b) some of the other possible implementations. For instance, 3D cubic data may choose to use each of its face as a macro patch. Also, a recent work~\cite{sun2019horizonnet} shows that horizontal tiles are naturally suitable for indoor layout task on panoramas, which points out that using horizontal tiles as macro patch in panorama generation may be an interesting future direction.}
    \end{figure}
    
    \vspace{2em}

\section{Beyond-Boundary Generation: More Examples and Details of Post-Training}
    \label{appendix:beyond-boundary-generation}

    \vspace{2em}

    \newlength{\oldintextsep}
    \setlength{\oldintextsep}{\intextsep}
    \setlength\intextsep{0pt}
    \begin{wrapfigure}{R}{0.3\linewidth}
        \includegraphics[width=\linewidth]{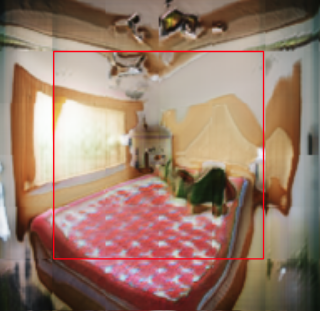}%
        \caption{Without \textit{\textbf{any extra}} training, original COCO-GAN can already perform slight extrapolations (\ie the edge of the bed extends out of the normal generation area annotated with the {\color{red}red} box), however, expectedly discontinuous on the edges.}
        \label{fig:extrapolation-without-training}
        \vspace{1em}
     \end{wrapfigure}
     
\begin{figure}[H]
        \vspace{2em}
        \centering
        \includegraphics[width=\linewidth]{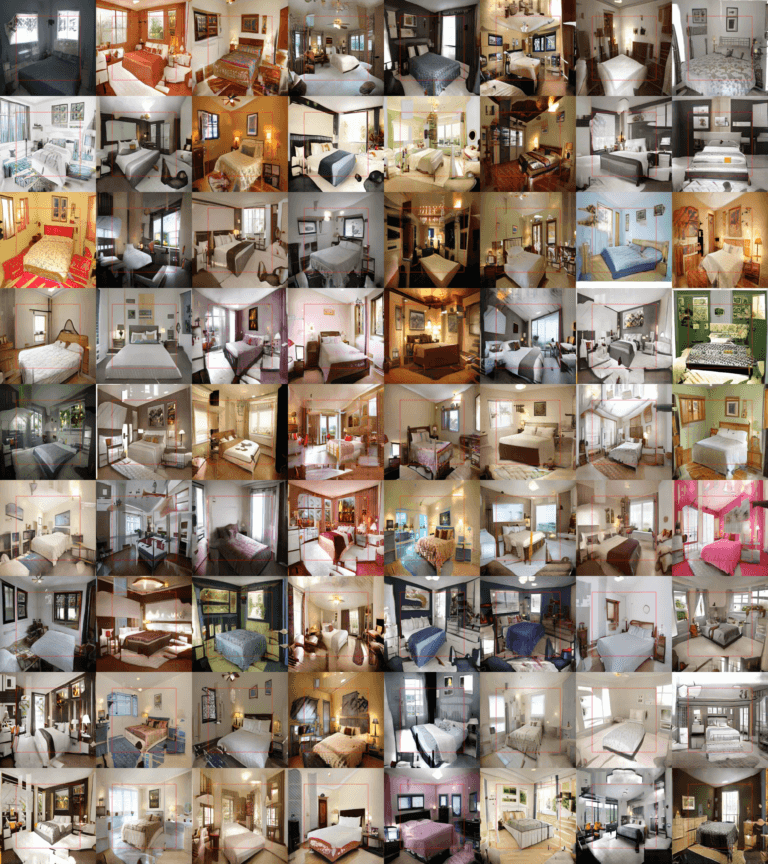}
        \caption{``Beyond-Boundary Generation'' generates additional contents by extrapolating the learned coordinate manifold. Note that the generated samples are $384\times384$ pixels, whereas \textbf{\textit{all}} of the training samples are of a smaller $256\times256$ resolution. The {\color{red} red} box annotates the $256\times256$ region for regular generation without extrapolation.}
        \label{fig:more-beyond-boundary-generation}
    \end{figure}

    We show more examples of ``Beyond-Boundary Generation'' in Figure~\ref{fig:more-beyond-boundary-generation}.

    Directly train with coordinates out of the $[-1,1]$ range (restricted by the real full images) is infeasible, since there is no real data at the coordinates outside of the boundary, thus the generator can exploit the discriminator easily. However, interestingly, we find extrapolating the coordinates of a manually trained COCO-GAN can already produce contents that seemingly extended from the edge of the generated full images (\eg, Figure~\ref{fig:extrapolation-without-training}). 
    
    With such an observation, we select to perform additional post-training on the checkpoint(s) of manually trained COCO-GAN (\eg, (N2,M2,S64) variant of \modelName that trained on LSUN dataset for 1 million steps with resolution $256\times256$ and a batch size 128). Aside from the original Adam optimizer that trains COCO-GAN with coordinates $\in \left[-1,1\right]$, we create another Adam optimizer with the default learning rate setup (\ie, $0.0004$ for $D$ and $0.0001$ for $G$). The additional optimizer trains COCO-GAN with additional coordinates along with the original coordinates. For instance, in our experiments, we extend an extra micro patch out of the image boundary, as a result, we train the model with $c''\in \left[-1.\overline{66},1.\overline{66}\right]$ (the distance between two consecutive micro patches is $2/(4-1)=0.\overline{66}$) and $c' \in [-2,2]$ (the distance between two consecutive macro patches is $2/((4-1)-1)=1$). We only use the new optimizer to train COCO-GAN until the discontinuity becomes patches becomes invisible. Note that we do not train the spatial prediction head $A$ with coordinates out of $\left[-1,1\right]$, since our original model has a tanh activation function on the output of $A$, which is impossible to produce predictions out of the range of $\left[-1,1\right]$.
    
    We empirically observe that by only training the first-two layers of the generator (while the whole discriminator at the same time) can largely stabilize the post-training process. Otherwise, the generator will start to produce weird and mottled artifacts. As the local textures are handled by later layers of the generator, we decide to freeze all the later layers and only train the first-two layers, which controls the most high-level representations. We flag the more detailed investigation on the root-cause of such an effect and other possible solutions as interesting future research direction.
    
\section{More Full Image Generation Examples}
    \label{appendix:more-full-images}
    \begin{figure}[H]
        \centering
        \subfloat[\small Selected generation samples.]{\includegraphics[width=\linewidth]{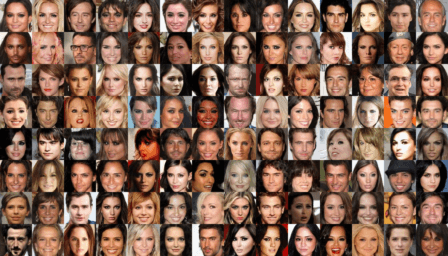}} \\ [0.5em]
        \subfloat[\small Random generation without calibration.]{\includegraphics[width=\linewidth]{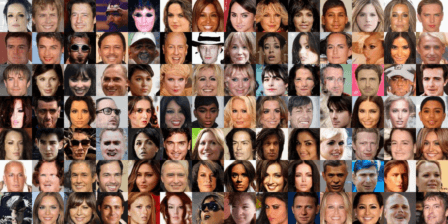}} \\ [0.5em]
        \subfloat[\small Generated micro patches.]{\includegraphics[width=\linewidth]{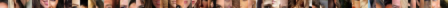}}
        \caption{Full images generated by \modelName on CelebA $128\times128$ with (N2,M2,S32) setting.}
        \label{fig:more-generation}
        \vspace{-1em}
    \end{figure}
    \blfootnote{Due to file size limit, all images are compressed, please access the full resolution pdf from: {\color{blue}\urlstyle{same}\url{https://goo.gl/5HLynv}}}
    
    \begin{figure}[H]
        \centering
        \subfloat[\small Selected generation samples.]{\includegraphics[width=\linewidth]{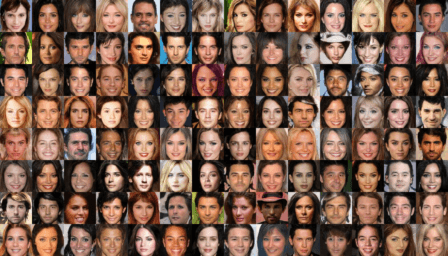}} \\
        \subfloat[\small Random generation without calibration.]{\includegraphics[width=\linewidth]{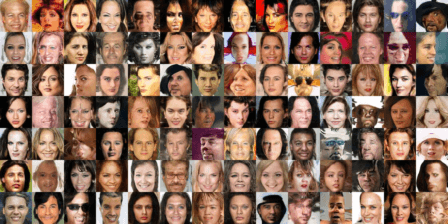}} \\
        \subfloat[\small Generated micro patches.]{\includegraphics[width=\linewidth]{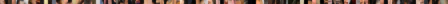}}
        \caption{Full images generated by \modelName on CelebA $128\times128$ with (N4,M4,S16) setting.}
    \end{figure}
    \blfootnote{Due to file size limit, all images are compressed, please access the full resolution pdf from: {\color{blue}\urlstyle{same}\url{https://goo.gl/5HLynv}}}
    
    \begin{figure}[H]
        \centering
        \subfloat[\small Selected generation samples.]{\includegraphics[width=\linewidth]{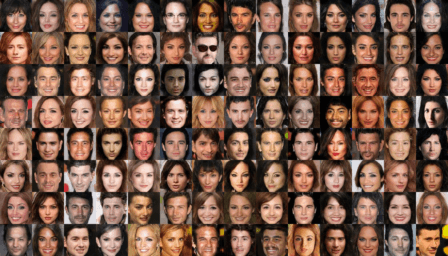}} \\
        \subfloat[\small Random generation without calibration.]{\includegraphics[width=\linewidth]{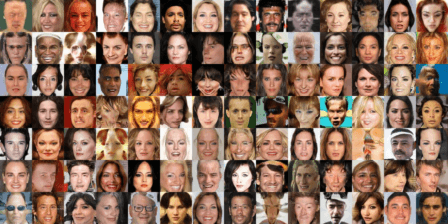}} \\
        \subfloat[\small Generated micro patches.]{\includegraphics[width=\linewidth]{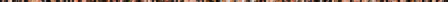}}
        \caption{Full images generated by \modelName on CelebA $128\times128$ with (N8,M8,S8) setting.}
    \end{figure}
    \blfootnote{Due to file size limit, all images are compressed, please access the full resolution pdf from: {\color{blue}\urlstyle{same}\url{https://goo.gl/5HLynv}}}
    
    \begin{figure}[H]
        \centering
        \subfloat[\small Selected generation samples.]{\includegraphics[width=\linewidth]{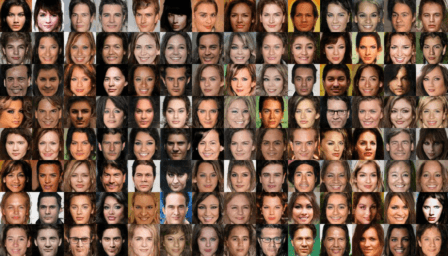}} \\
        \subfloat[\small Random generation without calibration.]{\includegraphics[width=\linewidth]{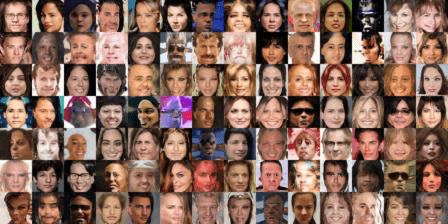}} \\
        \subfloat[\small Generated micro patches.]{\includegraphics[width=\linewidth]{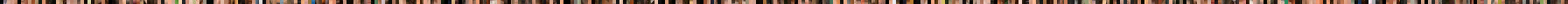}}
        \caption{Full images generated by \modelName on CelebA $128\times128$ with (N16,M16,S4) setting.}
    \end{figure}
    \blfootnote{Due to file size limit, all images are compressed, please access the full resolution pdf from: {\color{blue}\urlstyle{same}\url{https://goo.gl/5HLynv}}}
    
    \begin{figure}[H]
        \centering
        \subfloat[\small Selected generation samples.]{\includegraphics[width=\linewidth]{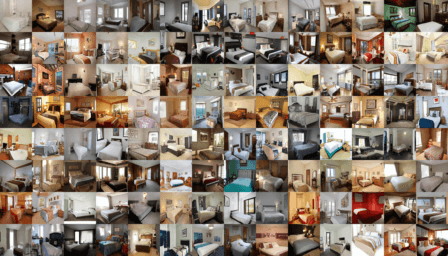}} \\
        \subfloat[\small Random generation without calibration.]{\includegraphics[width=\linewidth]{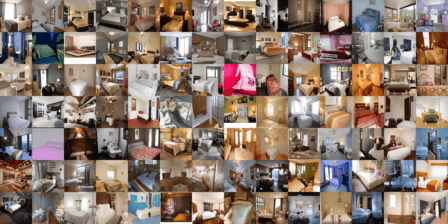}} \\
        \subfloat[\small Generated micro patches.]{\includegraphics[width=\linewidth]{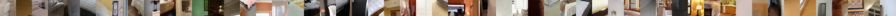}}
        \caption{Full images generated by \modelName on LSUN $256\times256$ with (N4,M4,S64) setting.}
    \end{figure}
    \blfootnote{Due to file size limit, all images are compressed, please access the full resolution pdf from: {\color{blue}\urlstyle{same}\url{https://goo.gl/5HLynv}}}

\section{More Interpolation Examples}

    \label{appendix:more-interp}
    \begin{figure}[H]
        \centering
        \subfloat[CelebA ($128\times 128$).]{
        \begin{tabular}{C{0.35\linewidth}}
            \textbf{Micro Patches Interpolation} \\ [0.5em]
            \includegraphics[width=\linewidth]{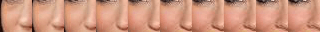} \\
            \includegraphics[width=\linewidth]{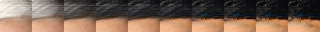} \\
            \includegraphics[width=\linewidth]{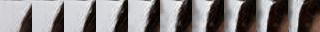} \\
            \includegraphics[width=\linewidth]{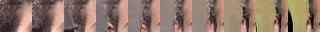} \\
            \includegraphics[width=\linewidth]{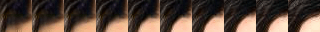} \\
            \includegraphics[width=\linewidth]{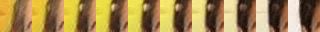} \\
            \includegraphics[width=\linewidth]{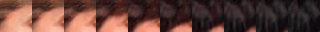} \\
            \includegraphics[width=\linewidth]{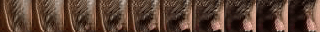} \\
            \includegraphics[width=\linewidth]{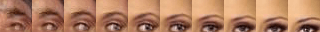} \\
            \includegraphics[width=\linewidth]{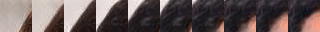} \\
            \includegraphics[width=\linewidth]{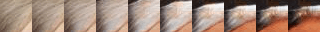} \\
            \includegraphics[width=\linewidth]{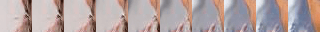} \\
        \end{tabular}%
        \begin{tabular}{C{0.35\linewidth}}
            \textbf{Full-Images Interpolation} \\ [0.5em]
            \includegraphics[width=\linewidth]{imgs/interp-full-image/CelebA/generated_images_interp_0.png} \\
            \includegraphics[width=\linewidth]{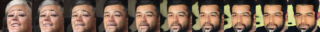} \\
            \includegraphics[width=\linewidth]{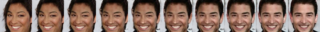} \\
            \includegraphics[width=\linewidth]{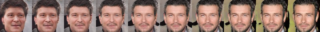} \\
            \includegraphics[width=\linewidth]{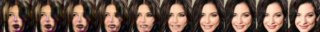} \\
            \includegraphics[width=\linewidth]{imgs/interp-full-image/CelebA/generated_images_interp_5.png} \\
            \includegraphics[width=\linewidth]{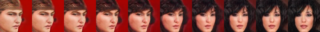} \\
            \includegraphics[width=\linewidth]{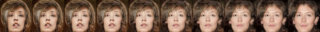} \\
            \includegraphics[width=\linewidth]{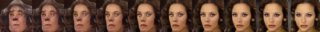} \\
            \includegraphics[width=\linewidth]{imgs/interp-full-image/CelebA/generated_images_interp_9.png} \\
            \includegraphics[width=\linewidth]{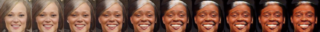} \\
            \includegraphics[width=\linewidth]{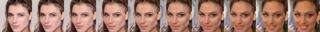} \\
        \end{tabular}}
        
        \subfloat[LSUN (bedroom category) ($256\times 256$).]{
        \begin{tabular}{C{0.35\linewidth}}
            \textbf{Micro Patches Interpolation} \\ [0.5em]
            \includegraphics[width=\linewidth]{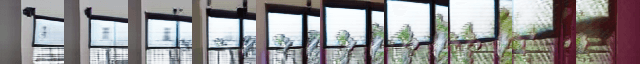} \\
            \includegraphics[width=\linewidth]{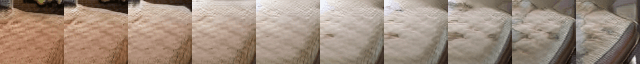} \\
            \includegraphics[width=\linewidth]{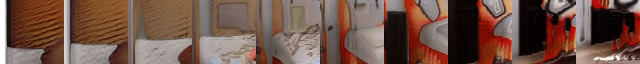} \\
            \includegraphics[width=\linewidth]{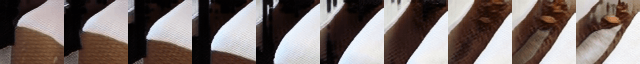} \\
            \includegraphics[width=\linewidth]{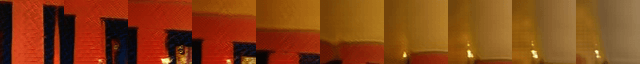} \\
            \includegraphics[width=\linewidth]{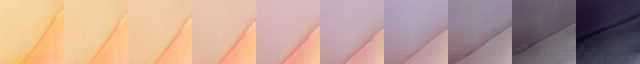} \\
            \includegraphics[width=\linewidth]{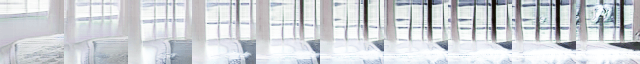} \\
            \includegraphics[width=\linewidth]{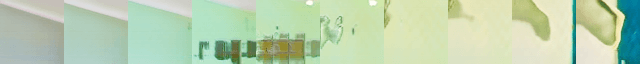} \\
            \includegraphics[width=\linewidth]{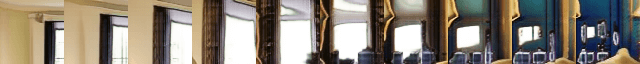} \\
            \includegraphics[width=\linewidth]{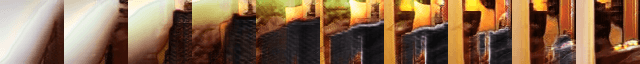} \\
            \includegraphics[width=\linewidth]{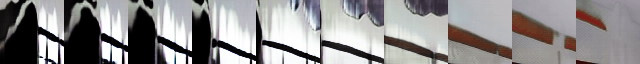} \\
            \includegraphics[width=\linewidth]{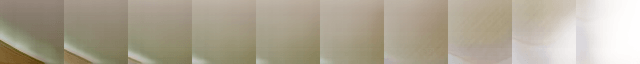} \\
        \end{tabular}%
        \begin{tabular}{C{0.35\linewidth}}
            \textbf{Full-Images Interpolation} \\ [0.5em]
            \includegraphics[width=\linewidth]{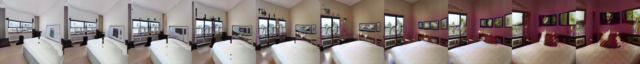} \\
            \includegraphics[width=\linewidth]{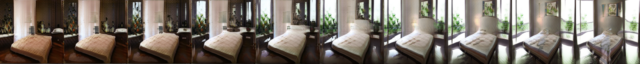} \\
            \includegraphics[width=\linewidth]{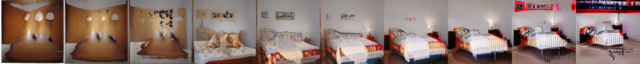} \\
            \includegraphics[width=\linewidth]{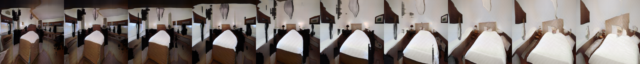} \\
            \includegraphics[width=\linewidth]{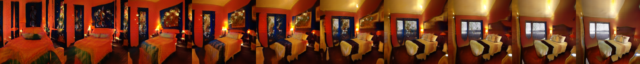} \\
            \includegraphics[width=\linewidth]{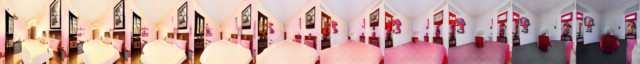} \\
            \includegraphics[width=\linewidth]{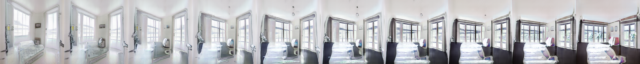} \\
            \includegraphics[width=\linewidth]{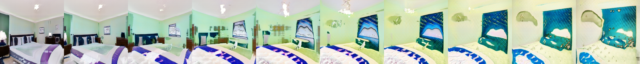} \\
            \includegraphics[width=\linewidth]{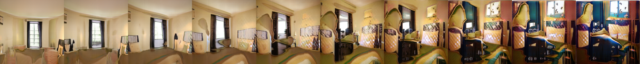} \\
            \includegraphics[width=\linewidth]{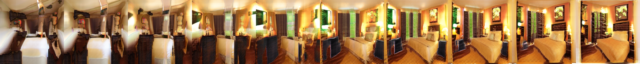} \\
            \includegraphics[width=\linewidth]{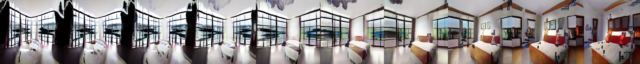} \\
            \includegraphics[width=\linewidth]{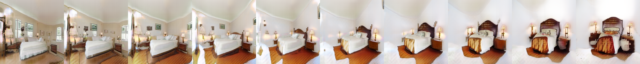} \\
        \end{tabular}}
        
        \caption{More interpolation examples. Given two latent vectors, \modelName generates the micro patches and full images that correspond to the interpolated latent vectors.}
    \end{figure}
    
    
\section{More Panorama Generation Samples}
    \label{appendix:panorama-samples}
    \begin{figure}[H]
        \includegraphics[width=\linewidth]{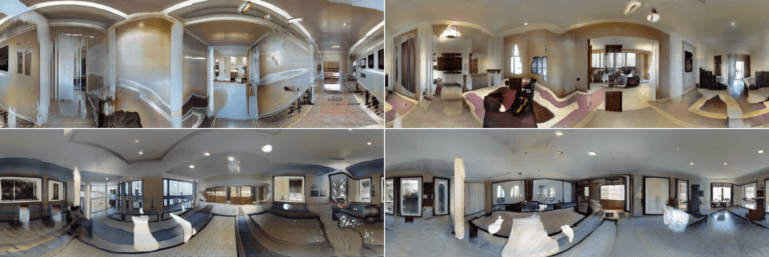}
        \caption{More examples of generated panoramas. All samples possess the cyclic property along the horizontal direction. Each sample is generated with a resolution of $768\times 256$ pixels, and micro patch size $64\times 64$ pixels.}
    \end{figure}
    
    \vspace{1em}
    
\section{Spatial Coordinates Interpolation}

    \label{appendix:interp-spatial}
    \begin{figure}[H]
        \begin{tabular}{cc|cc}
            \raisebox{-.3\height}{\includegraphics[width=0.3\linewidth]{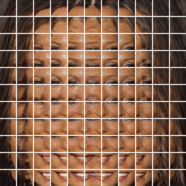}} & \includegraphics[width=0.1\linewidth]{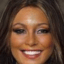} \hspace{1em} & \hspace{1em}
            \raisebox{-.3\height}{\includegraphics[width=0.3\linewidth]{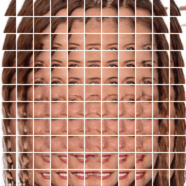}} & \includegraphics[width=0.1\linewidth]{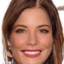} \\[5em]
            
            \raisebox{-.3\height}{\includegraphics[width=0.3\linewidth]{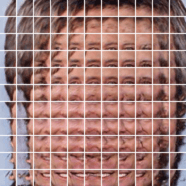}} & \includegraphics[width=0.1\linewidth]{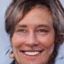} \hspace{1em} & \hspace{1em}
            \raisebox{-.3\height}{\includegraphics[width=0.3\linewidth]{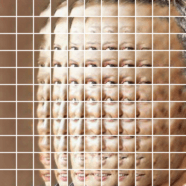}} & \includegraphics[width=0.1\linewidth]{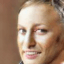} \hspace{1em}
        \end{tabular}
        \caption{Spatial interpolation shows the spatial continuity of the micro patches. The spatial conditions are interpolated between range $[-1, 1]$ of the micro coordinate with a fixed latent vector.}
    \end{figure}

\section{Ablation Study}
    \label{appendix:ablation}
    \vspace{1em}
    \begin{figure}[H]
        \centering
        \includegraphics[width=0.8\linewidth]{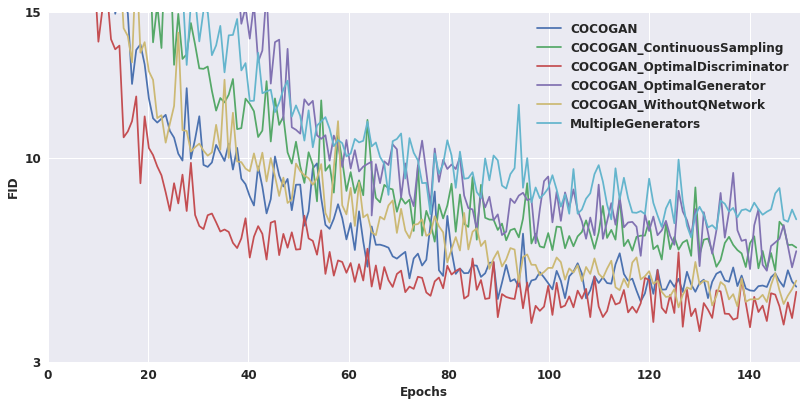}
        \caption{FID score curves of different variants of COCO-GAN in CelebA $64\times 64$ setting. Combined with Figure~\ref{fig:coco-family-samples}, the results do not show significant differences in quality between COCO-GAN variants. Therefore, COCO-GAN does not pay significant trade-off for the conditional coordinate property.}
        \label{fig:ablation}
        \vspace{1em}
    \end{figure}
    
    \begin{figure}[H]
        \centering
        \subfloat[COCO-GAN (ours).]{
            \includegraphics[trim={0 0 {0.26\textwidth} {0.26\textwidth}},clip,width=0.27\linewidth]{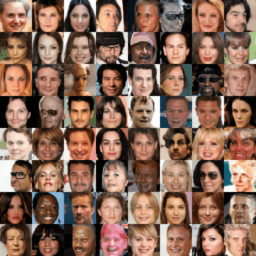}}%
        \hfill%
        \subfloat[COCO-GAN (cont sampling).]{
            \includegraphics[trim={0 0 {0.26\linewidth} {0.26\textwidth}},clip,width=0.27\linewidth]{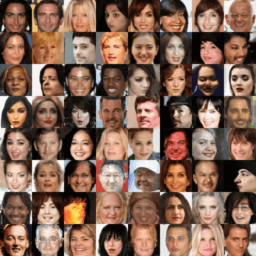}}%
        \hfill%
        \subfloat[COCO-GAN (optimal $D$).]{
            \includegraphics[trim={0 0 {0.26\linewidth} {0.26\textwidth}},clip,width=0.27\linewidth]{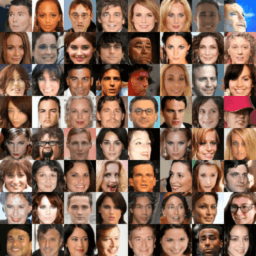}}
        
        \hfill%
        \subfloat[COCO-GAN (optimal $G$).]{
            \includegraphics[trim={0 0 {0.26\textwidth} {0.26\textwidth}},clip,width=0.27\linewidth]{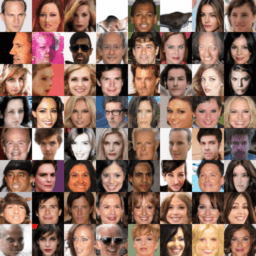}}%
        \hfill%
        \subfloat[Multiple generators.]{
            \includegraphics[trim={0 0 {0.26\linewidth} {0.26\textwidth}},clip,width=0.27\linewidth]{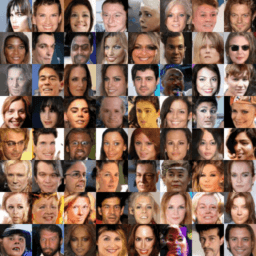}}%
        \hfill\hspace{0em}
        
        \caption{Some samples generated by different variants of COCO-GAN. Note that each set of samples is extracted at the epoch when each model variant reaches its lowest FID score. We also provide more samples for each of the variants at different epochs via following : {\color{blue}\urlstyle{same}\url{https://goo.gl/Wnrppf}}.}
        \label{fig:coco-family-samples}
        \vspace{-1em}
    \end{figure}

\section{Patch-Guided Image Generation}

    \label{appendix:patch-guided-image-generation}
    
    \begin{figure}[H]
        \centering
        \subfloat[(CelebA 128$\times$128) Real full images.]{\includegraphics[width=0.495\linewidth]{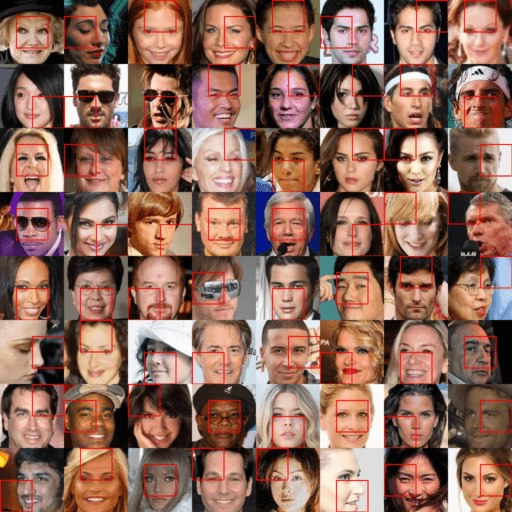}}%
        \hfill%
        \subfloat[(CelebA 128$\times$128) Real macro patches.]{\includegraphics[width=0.495\linewidth]{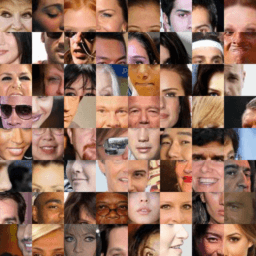}}
        
        \subfloat[(CelebA 128$\times$128) Patch-guided full image generation.]{\includegraphics[width=0.495\linewidth]{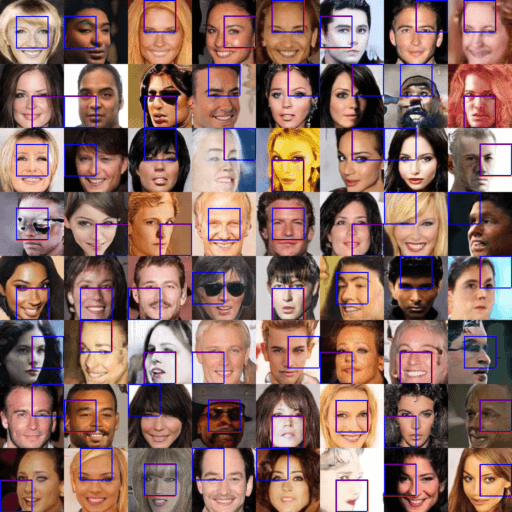}}
        \hfill%
        \subfloat[(CelebA 128$\times$128) Patch-guided macro patch generation.]{\includegraphics[width=0.495\linewidth]{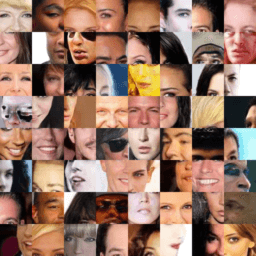}}
        
        \caption{Patch-guided image generation can loosely retain some local structure or global characteristic of the original image. (b) shows the patch-guided generated images based on $z_{\textit{est}}$ estimated from (a). The {\color{blue} blue} boxes visualize the predicted spatial coordinates $A(x')$, while the {\color{red} red} boxes indicates the ground truth coordinates $c'$. Since the information loss of cropping the macro patches from real images is critical, we do not expect (b) to be identical to the original real image. Instead, the area within blue boxes of (b) should be visually similar to (a), in the meanwhile, (b) should be globally coherent.}
    \end{figure}
    
\section{Training Indicators}
    \label{appendix:indicator-curves}
    \begin{figure}[H]
        \centering
        \subfloat[Wasserstein distance]{  
            \includegraphics[height=5cm]{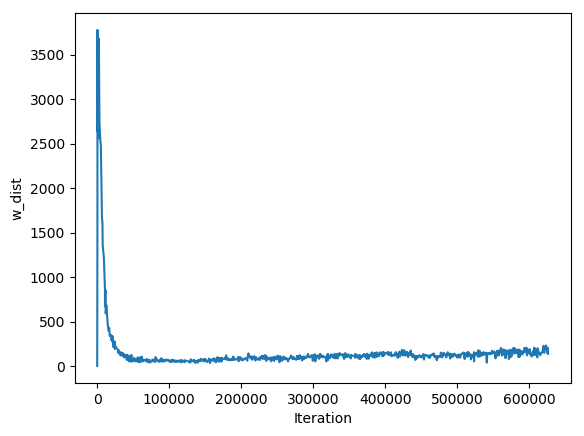}
        }
        \subfloat[FID]{
            \includegraphics[height=5cm]{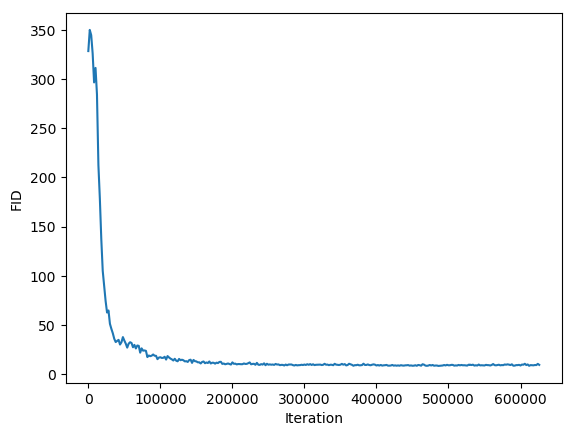}
        }
        
        \caption{Both Wasserstein distance and FID through time show that the  training of \modelName is stable. Both two figures are logged while training on CelebA with $128\times 128$ resolution.}
        \label{fig:indicator-curves}
    \end{figure}
    
    \vspace{1em}

\end{appendices}



\begin{multicols}{2}
{\small

}
\end{multicols}

\end{document}